\documentclass[8.5pt,twocolumn]{article}
\usepackage[super,sort&compress,comma]{natbib}
\usepackage{mhchem}
\usepackage{times,mathptmx}
\usepackage{sectsty}
\usepackage{balance} 

\usepackage{amsmath} \usepackage{lastpage}
\usepackage[format=plain,justification=raggedright,singlelinecheck=false,font=small,labelfont=bf,labelsep=space]{caption}

\usepackage{xifthen} % for conditional checks

\newif\ifproduction
\productionfalse  % Uncomment for normal mode (show figures)
\newcommand{\maybeincludegraphics}[2][]{%
  \IfFileExists{#2}{%
    \ifproduction
    \else
      \includegraphics[#1]{#2}%
    \fi
  }{%
  }%
}

\usepackage{listings}
\usepackage{helvet}

\usepackage{fancyhdr}
\usepackage{placeins} \usepackage{bm}

\usepackage[utf8]{inputenc}
\usepackage[T1]{fontenc}
\usepackage{ulem}
\usepackage{graphicx}
\usepackage{array}
\usepackage{fontawesome5} \usepackage{algorithm}
\usepackage[noend]{algpseudocode}
\usepackage{xcolor}
\usepackage{enumitem}
\usepackage{authblk}
 \usepackage{booktabs} 
 \usepackage{xspace}
 \definecolor{myblue}{RGB}{0, 102, 204} \usepackage[colorlinks=true, linkcolor=myblue, citecolor=myblue, urlcolor=myblue]{hyperref}
\usepackage{dblfloatfix} \usepackage{amssymb} \usepackage{marvosym} 

\usepackage{dblfloatfix}
\usepackage{subcaption}
\usepackage{nameref}  \newcommand{\appref}[1]{\hyperref[#1]{Appendix  \nameref{#1}}}

\newcommand{\supsecref}[1]{\hyperref[#1]{~\ref*{#1}}}
\newcommand{\supfigref}[1]{\hyperref[#1]{~\ref*{#1}}}
\newcommand{\suptabref}[1]{\hyperref[#1]{Supplementary Table~\ref*{#1}}}

\definecolor{BLUE}{named}{blue}
\definecolor{RED}{named}{red}

\newcommand{\revise}[2]{{#2}}

\newcommand{\update}[2]{{#2}}

\usepackage{titlesec}
\usepackage{titlecaps}

\titleformat{\section} {\normalfont\large\bfseries} {} {0pt} {}

\titleformat{\subsection} {\normalfont\normalsize\bfseries\sffamily} {} {0pt} {\titlecap} 

\titleformat{\subsubsection} {\normalfont\normalsize\bfseries\sffamily} {} {0pt} {\titlecap} 

\titlespacing{\section}{0pt}{12pt}{6pt} \titlespacing{\subsection}{0pt}{12pt}{6pt}
\titlespacing{\subsubsection}{0pt}{12pt}{6pt}

\usepackage{listings}
\definecolor{codegreen}{rgb}{0,0.6,0}
\definecolor{codegray}{rgb}{0.4,0.4,0.4}
\definecolor{codepurple}{rgb}{0.5,0,0.9}
\definecolor{backcolour}{rgb}{0.95,0.95,0.95}

\definecolor{lightgray}{rgb}{0.9,0.9,0.9}
\definecolor{lightpink}{rgb}{0.98,0.85,0.86}
\definecolor{lightblue}{rgb}{0.68,0.84,0.9}

\lstdefinestyle{mystyle}{
    backgroundcolor=\color{backcolour},   
    commentstyle=\color{codegreen},
    keywordstyle=\color{magenta},
    numberstyle=\tiny\color{codegray},
    stringstyle=\color{codepurple},
    basicstyle=\fontsize{7.5}{8}\selectfont\ttfamily\ttfamily,
    breakatwhitespace=false,         
    breaklines=true,
    breakindent=0pt,
    captionpos=b,                    
    keepspaces=true,                 
    numbers=none,                    
    numbersep=5pt,                  
    showspaces=false,                
    showstringspaces=false,
    showtabs=false,                  
    tabsize=2
}
\algnewcommand{\IfThenElse}[3]{\State \algorithmicif\ #1\ \algorithmicthen\ #2\ \algorithmicelse\ #3}
\algnewcommand{\IfThen}[2]{\State \algorithmicif\ #1\ \algorithmicthen\ #2}
\algrenewcommand\algorithmicrequire{\textbf{Input:}}
\algrenewcommand\algorithmicensure{\textbf{Output:}}

\newlist{tightenum}{enumerate}{1}
\setlist[tightenum]{label=\arabic*., noitemsep, topsep=0pt, parsep=0pt, partopsep=0pt}
\newlist{tightitem}{itemize}{1}
\setlist[tightitem]{label=\textbullet, noitemsep, topsep=0pt, parsep=0pt, partopsep=0pt}

\usepackage{amssymb}
\usepackage[utf8]{inputenc}
\usepackage{tipa} 

\definecolor{myblue}{rgb}{0.0, 0.2, 0.6}  \definecolor{myred}{rgb}{0.8, 0.0, 0.0}   
\def\ourmodel{\textsc{\textsf{{\textcolor{black}{Matter}\textcolor{black}{iX}}}}\xspace}

\title{\textbf{\ourmodel: Towards a Digital Twin for Robotics-Assisted Chemistry Lab Automation}}
\date{}

\usepackage{eso-pic}
\author[1,2,3,*]{Kourosh Darvish}
\author[1]{Arjun Sohal}
\author[1]{Abhijoy Mandal}
\author[4]{Hatem Fakhruldeen}
\author[4]{Nikola Radulov}
\author[4]{Zhengxue Zhou}
\author[4]{Satheeshkumar Veeramani}
\author[1]{Joshua Choi}
\author[1]{Sijie Han}
\author[1]{Brayden Zhang}
\author[1]{Jeeyeoun Chae}
\author[4]{Alex Wright}
\author[1,2]{Yijie Wang}
\author[5]{Hossein Darvish}
\author[1,3]{Yuchi Zhao}
\author[1,3]{Gary Tom}
\author[1,2]{Han Hao}
\author[1,3]{Miroslav Bogdanovic}
\author[4]{Gabriella Pizzuto}
\author[4]{Andrew I. Cooper}
\author[1,2,3,6,7]{Alán Aspuru Guzik}
\author[1,3]{Florian Shkurti}
\author[8]{Animesh Garg}

\affil[1]{University of Toronto, Canada}
\affil[2]{Acceleration Consortium, Canada}
\affil[3]{Vector Institute, Canada}
\affil[4]{University of Liverpool, UK}
\affil[5]{University of Salento, Italy}
\affil[6]{NVIDIA}
\affil[7]{Canadian Institute for Advanced Research}
\affil[8]{Georgia Institute of Technology, USA}

\affil[*]{\texttt{kourosh.darvish@utoronto.ca}}

\usepackage{abstract}  

\definecolor{darkblue}{RGB}{0, 35, 130}

\AddToShipoutPictureBG*{\ifnum\value{page}=1
    \AtPageUpperLeft{\put(\dimexpr0.5\paperwidth-0.525\textwidth\relax,-30){\color{darkblue}\rule{\textwidth}{3pt}  }}\fi
}

\begin{document}

\twocolumn[{\vspace{0mm}
  \textbf{Article}
  \vspace{2mm}
  \hrule height 0.4pt\relax \vspace{-7mm}
  \maketitle
    \hspace*{\absleftindent}\rule{\dimexpr\textwidth - \absleftindent - \absrightindent\relax}{0.4pt} \vspace{-5mm}
  \begin{abstract}
  \fontsize{11.5}{13.5}\selectfont \vspace{-5mm}
\noindent
% \href{https://ac-rad.github.io/matterix}{\texttt{https://ac-rad.github.io/matterix}}\\
\textbf{\update{}{Abstract.}} Accelerated materials discovery is critical for addressing global challenges. However, developing new laboratory workflows relies heavily on real-world experimental trials, and this can hinder scalability because of the need for numerous physical make-and-test iterations.
\update{}{Here,} we present \ourmodel, a multi-scale, GPU-accelerated robotic simulation framework designed to create high-fidelity digital twins of chemistry labs, thus accelerating workflow development.
This multi-scale digital twin simulates robotic physical manipulation, powder and liquid dynamics, device functionalities, heat transfer, and basic chemical reaction kinetics.
This is enabled by integrating realistic physics simulation and photorealistic rendering with a modular GPU-accelerated semantics engine, which models logical states and continuous behaviors to simulate chemistry workflows across different levels of abstraction.
\ourmodel streamlines the creation of digital twin environments through \update{extensive}{} open-source asset libraries and interfaces, while enabling flexible workflow design via hierarchical plan definition and a modular skill library that incorporates learning-based methods. 
Our approach demonstrates sim-to-real transfer in robotic chemistry setups, reducing reliance on costly real-world experiments and enabling the testing of hypothetical automated workflows in silico.
\update{The accompanying open-source libraries empower researchers and lab automation professionals to develop and validate innovative experimental protocols rapidly.}{}
\end{abstract}
}]

\section{\update{}{Introduction}}
\label{sec:introduction}
\noindent
The accelerated discovery of new materials is fundamental to addressing global challenges such as resilient and flexible pharmaceutical manufacturing, and achieving net zero targets~\cite{stein2019progress}. The urgency of these challenges demands a paradigm shift in the way that we approach materials discovery.
However, materials research remains heavily reliant on manual design and execution of experimental workflows. This is slow and costly because of the inherent challenges of automating complex experimental workflows with a high degree of variability. Recent advances in laboratory automation and Self-Driving Labs (SDLs) have led to the deployment of a diverse array of robotic systems to accelerate the discovery of new materials~\cite{Tom2024}. 
These systems have been instrumental in automating workflows ranging from solubility screening~\cite{shiri2021Solubility} and synthesis~\cite{coley2019robotic} to electrochemistry~\cite{darvish2024organa}, catalysis~\cite{Burger2020}, solid-state chemistry~\cite{Lunt2024}, chirality detection in nanocrystals~\cite{Li2020}, and the exploration of polymer electronics~\cite{Vriza2023}.

% \begin{figure*}[!t]
%     \centering
%     \maketitle
%     \includegraphics[width=\textwidth]{Images/Fig1_overview.pdf}
%     \captionof{figure}{}
% \label{fig:architecture}
% \end{figure*}

\begin{figure*}[!t]
    \centering
    \maketitle
    \maybeincludegraphics[width=0.95\textwidth]{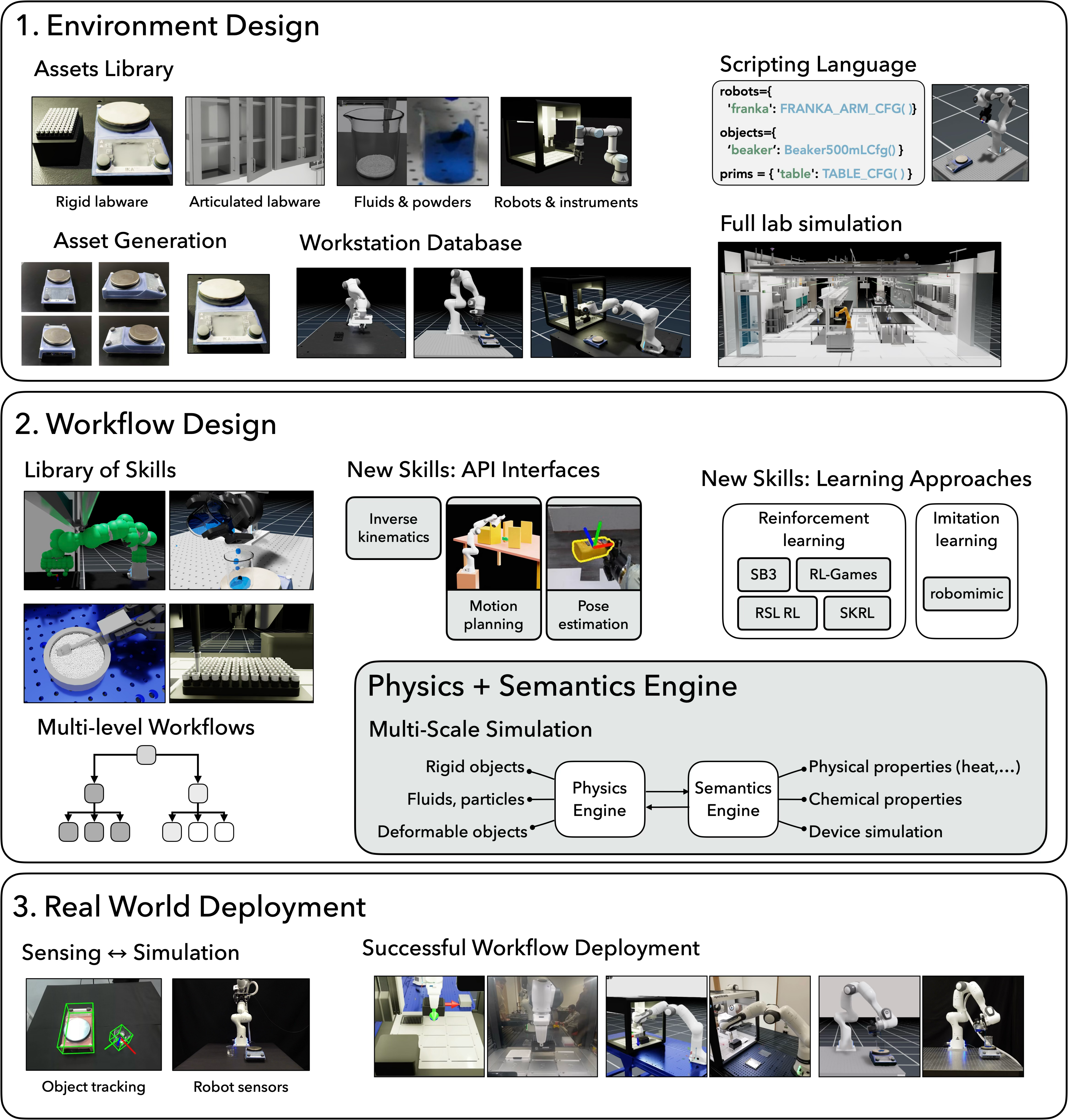}
    \captionof{figure}{\textbf{\ourmodel architecture and components.} 
\ourmodel generates a digital twin environment by specifying objects and their poses using a library of digital and photorealistic assets (1).
It solves a given workflow to perform the chemistry experiment (2). To perform the experiment, it trains new policies that are not available in the skill library but are required for experimenting. A multi-step workflow is generated by a human or large language model. Using a database of known reactions and chemical kinetics approximation, \ourmodel verifies if the target chemical is synthesized.
\ourmodel is deployed in the real world and performs a chemistry experiment (3). The object poses are perceived using a camera stream in the real environment, and the simulated robots are matched with the real robots.}
\label{fig:architecture}
\end{figure*}

Most, if not all, autonomous material discovery experiments today are designed, developed, and deployed directly in real-world environments.
This approach is complex and time-consuming, encompassing algorithmic development, hardware interfacing, and software integration, making it a primary bottleneck in developing fully autonomous self-driving labs~\cite{pelkie2025democratizing}.
For example, it took us three years to build the follow-up to our original 'mobile robotic chemist'~\cite{Burger2020}, which was an autonomous robotic workflow for
solid-state chemistry~\cite{Lunt2024}. This was done entirely in the laboratory, without any simulations or digital twins.
This approach both limits testing efficiency and impedes the progress of learning-based methods for robotic chemists, necessitating large amounts of data that are both costly to collect from real-world setups. 
There are inherent hurdles in integrating dexterous robots, automated platforms, and multi-robot system integration within the dynamic of an unpredictable mixed human/robot chemistry lab.
As a result, the majority of robotic systems implemented in chemistry laboratories today are limited to relatively simple experimental workflows. 
The effective handling and interpretation of multi-modal data generated during these experiments poses another considerable challenge~\cite{Tom2024}.
To address these challenges, simulators offer a critical solution to these challenges by providing a safe and efficient virtual environment for development and testing.
These platforms enable rapid iteration of robotic planning and control algorithms, sensor integration, and experimental protocols without the constraints and risks associated with real lab setups. 
By generating synthetic data, simulators can accelerate the training of learning-based models, reducing the reliance on costly and time-consuming real-world experiments.
By enabling granular, hardware-aware workflow descriptions that specify robotic actions and environmental conditions with high precision, simulators enhance experimental reproducibility, complementing widely adopted hardware-agnostic standards like $~\chi$DL~\cite{Mehr2020}.

Digital twin technologies~\cite{tao2024advancements} have found successful applications across domains such as manufacturing and autonomous systems, enabling process optimization, predictive maintenance, and synthetic data generation through real-time monitoring and virtual-physical synchronization~\cite{lu2020digital, almeaibed2021digital}. In self-driving cars, for instance, digital twins enhance safety validation and decision-making by simulating complex scenarios~\cite{li2024choose}. Beyond industrial and automotive contexts, they also show transformative potential in accelerating scientific discovery and personalized medicine~\cite{laubenbacher2024digital}. Despite these advances, the adaptation of digital twins to self-driving labs—where they could address critical challenges in lab automation, reproducibility, and experimental design—remains underexplored.

\begin{figure*}[!t]
    \centering
    \maketitle
\maybeincludegraphics[width=\textwidth, trim=0mm 35mm 0mm 35mm, clip]{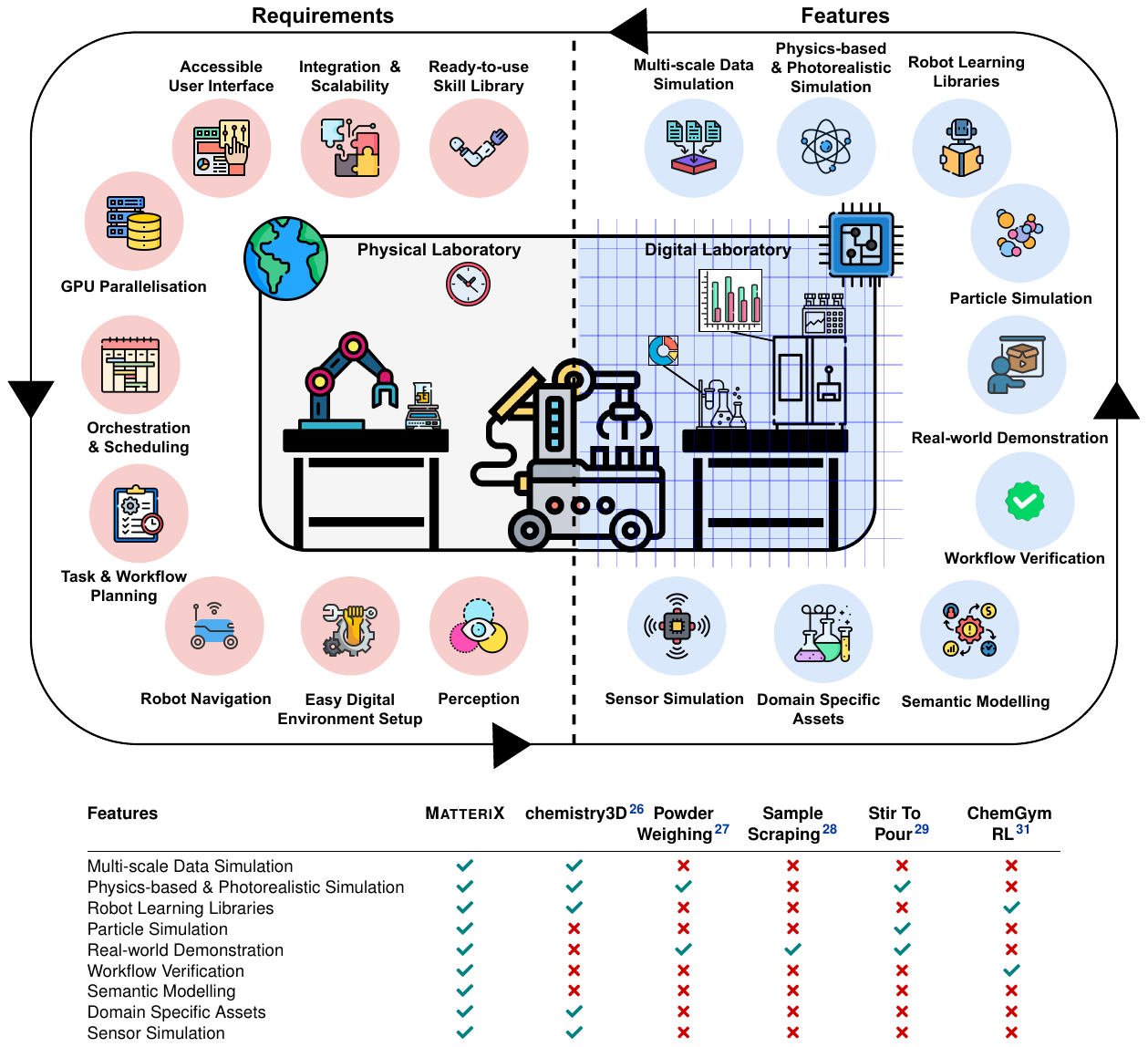}
    \captionof{figure}{\textbf{Digital twin requirements and comparison with \ourmodel.} The top figure illustrates key requirements and benefits of digital twins in chemistry lab automation, while the bottom compares \ourmodel with related works.
}
\label{fig:requirements}
\end{figure*}

For this purpose, we propose \ourmodel. This is a multi-scale robotic simulation framework (\autoref{fig:architecture}) for developing digital twins of laboratories, to facilitate the design and evaluation of experimental workflows, and to allow easy deployment to the real environment. 
\ourmodel is a comprehensive set of open-source 
tools building on the Nvidia Isaac Sim physics engine\cite{IsaacSim}
and Isaac Lab framework~\cite{Mittal2023Orbit}.
Because it is based on Isaac Sim and Isaac Lab, \ourmodel comes with inherently fast and accurate physics simulation, parallelized GPU simulation, contact-rich interactions (\update{e.g.,}{like} Signed Distance Field (SDF)-based collision meshes for precise collision detection~\cite{Narang2022FACTORY} and convex decomposition for simplified collision shapes~\cite{Mamou2016ConvDec}), Application Programming Interfaces (APIs) for object definition, and a range of robot learning libraries (Stable Baselines 3~\cite{Raffin2021SB3}, robomimic~\cite{Mandlekar2021}, RL-Games~\cite{Makoviichuk2021RLGames}, RSL RL~\cite{Rudin2022}, and SKRL~\cite{serrano2023} among others).
We further extend the framework with several features, including
a \textit{semantics engine}; that is, a system enabling the modular simulation of behaviours, such as chemical reaction kinetics, heat transfer, and logical processes (which are not inherently supported by robotics simulators).
Through these features, \ourmodel addresses the current limitations of the suite of tools available for chemistry lab automation as showcased in \autoref{fig:requirements}, which highlights the requirements from a digital twin of a chemistry lab and compares \ourmodel with the recent works in this domain.
The advantage of using \ourmodel is it adds many of the key features required for chemistry lab automation, including particle simulation, simulation-to-reality deployment in real labs and workflow verification which is imperative in safety-critical environments as is our domain application within a single framework.
Simulation environments, such as those detailed in~\cite{Li2024chemistry3droboticinteractionbenchmark, Kadokawa2023PowderWeighing, Pizzuto2024Scraping, LopezGuevara2020StirToPour, vescovi2023towards} showcase the role that physics-based simulations can play for accurately modelling environments used for robotic chemist skill learning and developing better perception systems.
Other simulators are more tailored for developing and validating automated synthesis workflows or automating research operations, \update{as highlighted in}{and} recent works~\cite{Beeler2024chemgymrl,rihm2024digital} underscore their role in streamlining material discovery.

% \begin{figure*}[!t]
%     \centering
%     \maketitle
% \includegraphics[width=\textwidth, trim=0mm 35mm 0mm 35mm, clip]{Images/Fig2_ DigitalTwin_Features_Needs_Literature.pdf}
%     \caption{}
% \label{fig:requirements}
% \end{figure*}

~\ourmodel aims to be a comprehensive tool that embeds key features for automating chemistry experiments towards empowering automation chemists to embrace digital tools for accelerating the design and deployment of automated experiments.
It is also aimed to be a tool for collecting extensive datasets required for advancing robotic manipulation, refining control algorithms, and optimising experimental workflows, all within a highly controllable and efficient setting.

In particular, our contributions are the following: 

\textbf{1) Multi-scale and GPU-based chemistry experiment workflow simulation.} An open-source simulator that can simulate chemistry lab experiments. 
\ourmodel extends NVIDIA Isaac Lab~\cite{Mittal2023Orbit}, the physics engine of NVIDIA Isaac Sim, and its particle-based simulation with a semantics engine.
The multi-scale nature of \ourmodel allows the design and verification of full chemical experiment workflows. 

\textbf{2) Easy generation of new chemistry lab environments.} \ourmodel allows users to easily design new environments by identifying the set of assets and their locations (poses) in the digital twin environment.
It comes with an extensive library of digital and photo realistic assets.
These assets are either curated from open-source resources or custom-built using \update{CAD design}{Computer-Aided Design (CAD)} and 3D scene reconstruction techniques. This includes diverse glassware, (mobile) robotic manipulators, automated platforms, and other essential laboratory equipment, facilitating the easy setup of detailed simulation environments. 
Moreover, it allows users to generate digital twin environments based on real environments using 3D perception and scanning methods, interfaced with lab devices and robots.

\textbf{3) Design of workflows for flexible chemistry experiments.} \ourmodel enables users to easily build chemistry workflows. This has been enabled using a tree-structure hierarchical state machine and a library of skills that allows users to combine those skills to control multiple automation devices and robots (agents) in the digital twin environments. These skills are based on several methods such as reinforcement and imitation learning, classical motion planners with/without obstacle avoidance, and whole-body mobile robot controllers.
The combination of hierarchical state-machine-based and learning-based methods to demonstrate policies for various lab tasks facilitates the reuse of learned skills in novel experimental workflows. The combination of computer vision and safe motion planners with obstacle avoidance allows users to transfer samples and objects flexibly without the need for hard-coded robot trajectories.

\textbf{4) Deployment of workflow solutions to real laboratory setups.} \update{The first real-world demonstration of}{We demonstrate} sim-to-real transfer for autonomous chemistry labs, deploying general-purpose robots and automated platforms programmed using \ourmodel in a real-world laboratory setting. These experiments validate the repeatability and transferability of our approach, paving the way for broader lab automation.

%  \begin{figure*}[!t]
%     \centering
% \includegraphics[width=0.85\textwidth]{Images/Fig3_fulllab_workstations.pdf}
%     \captionof{figure}{}
% \label{fig:lab_environments}
% \end{figure*}

\begin{figure*}[!t]
    \centering
\maybeincludegraphics[width=0.85\textwidth]{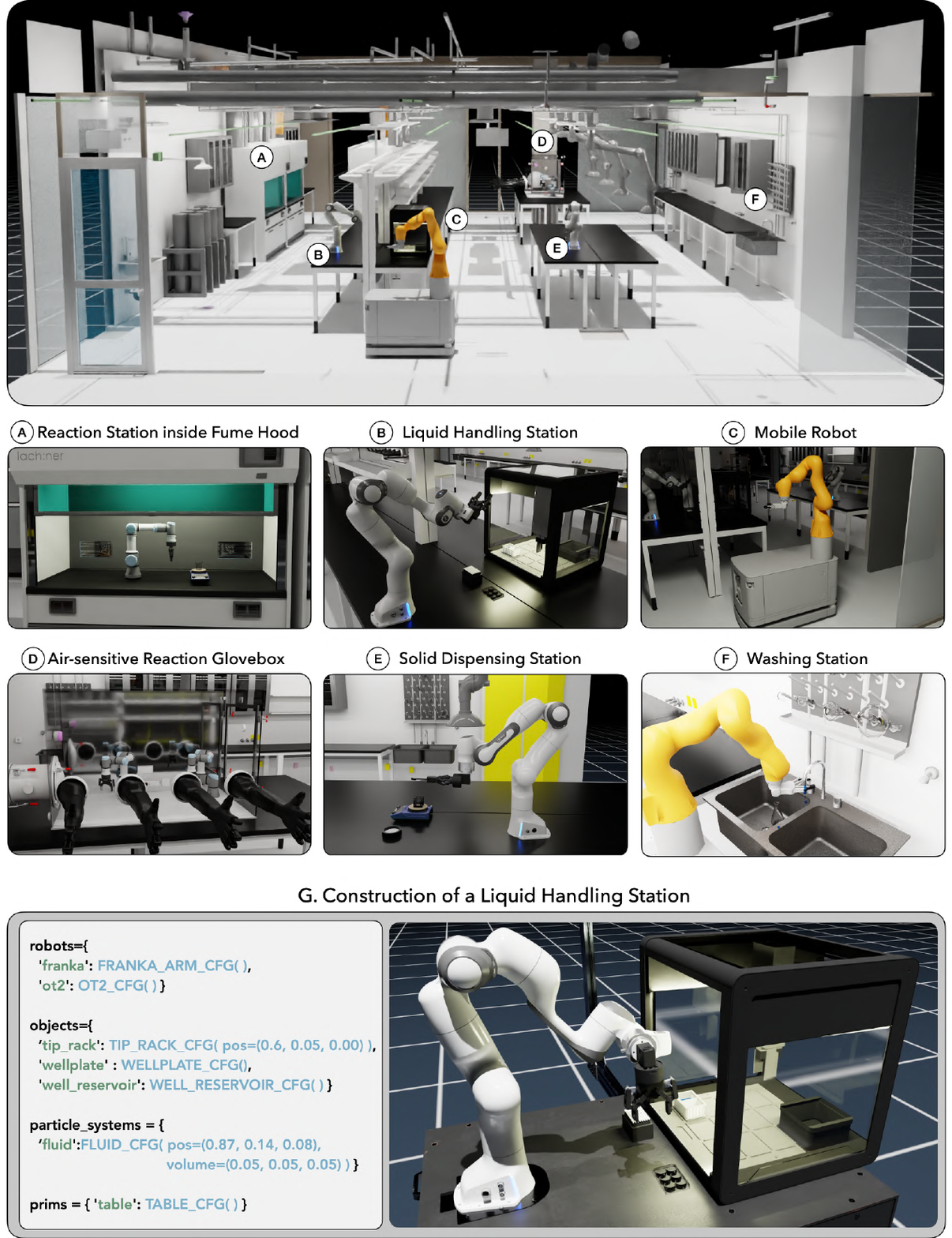}
    \captionof{figure}{\textbf{Examples of \ourmodel digital twin full chemistry laboratory environment and its workstations.} A chemistry lab simulation framework comprising different (mobile) robotic chemists, instruments, and glassware across various laboratory tasks, including materials and fluids (A-F). (G) demonstrates how an environment is created.
}
\label{fig:lab_environments}
\end{figure*}

\section{Results}\label{sec:results}

We assess the performance of different components of \ourmodel, including the generation of digital environments and a full self-driving lab setup, skill library, and semantics engine. Following that, we demonstrate how \ourmodel simulates two different multi-step chemistry experiment workflows (one organic, one inorganic).
Additionally, we demonstrate and evaluate the deployment of our method from simulation to a real-world setup, showcasing its ability to execute a long-horizon, contact-rich chemistry experiment with a single robot arm and to conduct multi-system experiments using a robot arm and an automated liquid handler.

\subsection{Virtual Environment Generation Results}

\ourmodel provides an open-source library of 3D assets that are commonly found in chemistry labs. 
This consists of \textbf{2} pieces of lab equipment for contained experiments (glove box, fumehood), \textbf{2} fixed-base robot arms (7 degrees of freedom (DoF) Franka Emika Panda and 6 DoF UR3e), \textbf{1} mobile manipulator (KUKA KMR iiwa), \textbf{1} bench-top liquid handler (Opentrons (OT-2)),  \textbf{12} different lab objects (beakers, pipettes, graduated cylinders, spatulas, well reservoirs, well plates, glass bottles, flasks, funnels, racks, test tubes, vials), \textbf{2} lab instruments (stirring/heating plate, scale and syringe pump), \textbf{2} lab tables and \textbf{1} full 3D model of a lab environment.
This list is not exhaustive and users can extend it by adding their own assets.
\ourmodel also enables the addition of new assets using Neural Radiance Fields (NeRF), a 3D computer vision technique for scene reconstruction that allows the synthesis of photorealistic novel views~\cite{mildenhall2021nerf}.
\autoref{fig:architecture} illustrates a virtual asset of a lab instrument created by the NeRF model, using an object-centric video provided by the user.
This approach is particularly useful for constructing a USD representation of an object from a real-world setup when the corresponding asset is absent in the asset library. The physical properties of the generated asset\update{ (\textit{e.g.,} mass,  friction coefficient)}{, like mass or friction,} are manually specified by the user. Future research could explore learning methods and physical robot interactions with objects to automate or refine this process~\cite{zeng2020tossingbot, xu2019densephysnet}.
\ourmodel also supports the definition of multiple, heterogeneous embodied agents (robots) in an environment, facilitating their collaborative execution of complex autonomous laboratory experiments.

All these assets are modelled by Universal Scene Description (USD)~\cite{openusd_software_2025}. 
USD is a high-performance, extensible platform for collaborative 3D scene construction, enabling robust asset interchange, composition, and scalability across animation, visual effects, and digital content creation. 
\update{The assets have collision with physical (\textit{e.g., }mass, collision, friction) and textual representations (visual appearance, \textit{e.g.,} color patterns, roughness, or bump maps),  object hierarchies, transformations (position, rotation, scale), material properties (surface interaction with light, \textit{e.g.,} shaders, reflectivity), and metadata such as object names, labels, or other descriptive information.}{Assets specify physical properties (mass, friction, contact parameters), textural/visual representations (color patterns, roughness, bump maps), object hierarchies, transformations (position, rotation, scale), material properties (interaction with light, shaders, reflectivity), and metadata (object names, labels, and other descriptive information).}
Moreover, the user can attach additional properties to an asset in the asset library and query them during the simulation. 
For example, the user can specify a list of object key points (\update{\textit{e.g.,} }{}place to grasp, location to perform insertion, etc.) and use that information for solving automation tasks.

Fluids and powders are fundamental components in chemistry labs - most chemicals, other than gases, tend to be either fluids or powders. 
\ourmodel provides users with the capability to easily incorporate fluids and granular media into their chemistry workstation designs, enabling more realistic physical simulations of their experiments. \revise{}{To add liquid or powder in the simulation, users can specify the quantity and location for initialization. Particle properties, such as physical behavior or appearance, can also be customized beyond the default settings.}
They are simulated using a GPU-accelerated position-based-dynamics (PDB) particle simulation~\cite{macklin2013position}, allowing for fast and accurate computation and accounting for their interaction with all other objects in the scene.

\paragraph{Workstation Virtual Environments}
To create and initialise a laboratory workstation, the users can add the assets from the asset library to a configuration file, which \ourmodel would then use to generate a virtual environment.
To achieve this, one simply needs to create a dictionary, with the asset naming and typing.
\autoref{fig:lab_environments} demonstrates different chemistry lab workstations in \ourmodel. 
These environments enable users to design, develop, and deploy modular robotic and chemistry workstations towards advancing the adoption of self-driving labs.
While simpler environments are well-suited for training robot policies, more complex environments, which incorporate a wider range of lab equipment and realistic layouts, are essential for conducting thorough feasibility studies, optimising lab layout designs for efficiency, and implementing comprehensive monitoring systems for automated experiments.
A full list of environments and assets and their functionalities are provided in~\supsecref{sec:appendix:assets-description} and ~\supsecref{sec:appendix:enviornments-tasks}.

\autoref{fig:lab_environments}G illustrates an example of a liquid handling station and the script to initialise the environment. 
The environment would comprise different robots; \update{e.g.}{for example}, a Franka Panda arm equipped with a Robotiq 2F-85 adaptive gripper and an OpenTrons OT-2 bench-top liquid handler. 
Several key objects need to also be defined, including an OpenTrons pipette tip rack, a 6-well wellplate, an IKA RET control scale, and a two-well reservoir.  
The specification of locations of numerous objects in chemistry labs can be challenging, to address this challenge, ~\ourmodel provides a utility function that allows users to initialize environments with a USD with several objects.
For example, in the liquid handling station, the 96 pipette tip with the rack is defined as a nested multi-object USD.

\paragraph{Full Self-driving Lab Environment}
The digital twin of a full chemistry lab is shown in \autoref{fig:lab_environments}.
It is modeled after one of the Acceleration Consortium’s inorganic self-driving labs (SDLs) at the University of Toronto.
The lab features \textbf{4} fume hoods, \textbf{1} benchtop liquid handling station, \textbf{1} solid dispensing station, \textbf{1} glove box for air- and moisture-sensitive reactions, and \textbf{1} washing and drying station.
Robotic systems include \textbf{5} fixed-base manipulators (Franka Emika Panda and Universal Robots) and \textbf{1} mobile manipulator (KUKA KMR iiwa).
Users can define and execute long-horizon tasks and workflows within this digital twin environment.

% \begin{figure*}[!h]
%     \centering
%     \includegraphics[width=0.8\textwidth, trim=0mm 0mm 0mm 0mm, clip]{Images/Fig4_workflows.pdf}
%     \captionof{figure}{}
% \label{fig:fig_skills}
% \end{figure*}

\begin{figure*}[!t]
    \centering
    \maybeincludegraphics[width=0.8\textwidth, trim=0mm 0mm 0mm 0mm, clip]{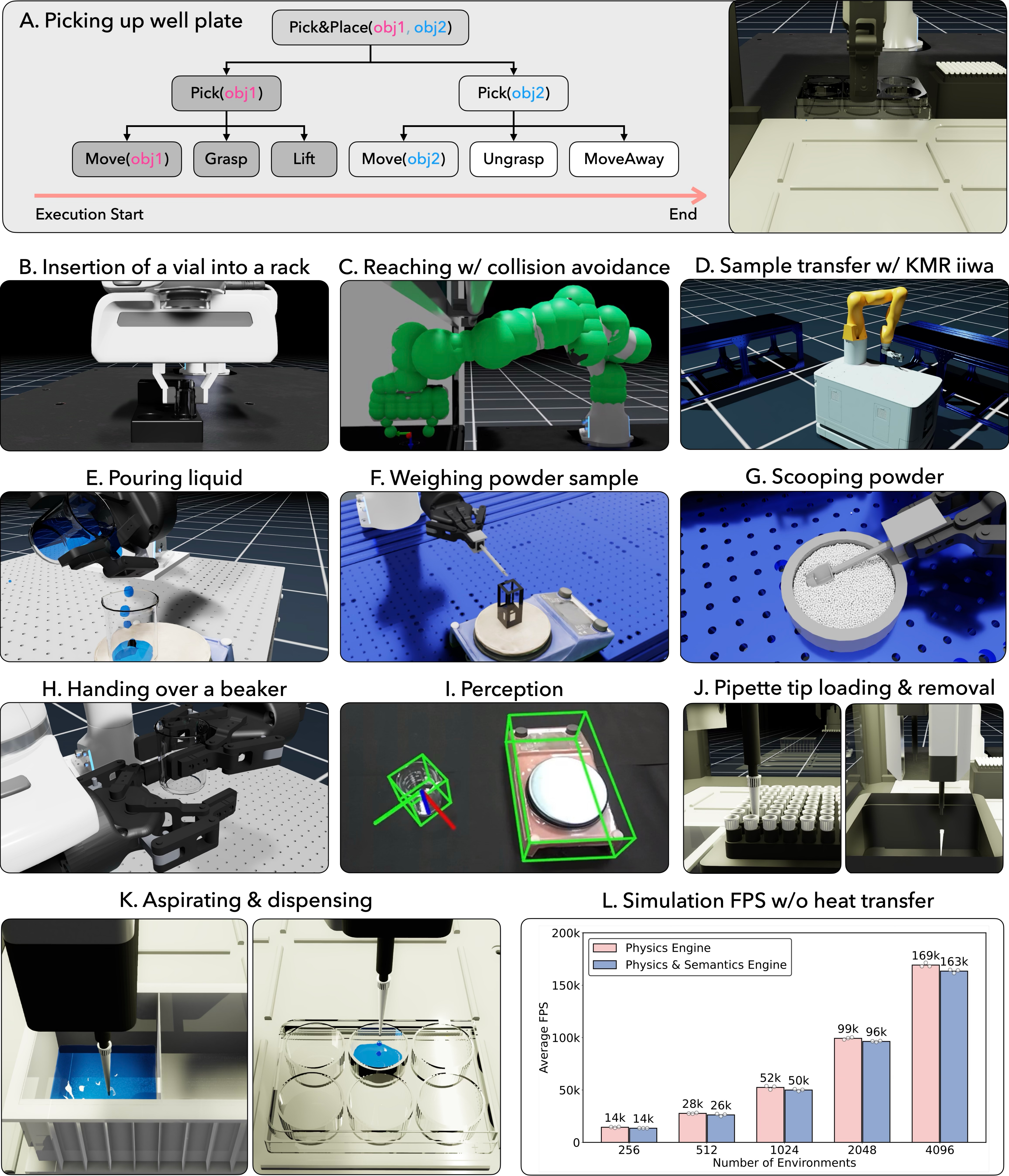}
    \captionof{figure}{\textbf{A library of manipulation, perception skills, and device functionalities in \ourmodel.} The figure demonstrates the behaviour and performance of our integrated physics and semantics engine. (A) illustrates the creation of abstract skills using a hierarchical state machine. (B–F) showcase various skills for manipulating rigid and granular objects. (I) presents object pose estimation via computer vision. (J–K) highlight liquid-handling device functionalities enabled by the integrated semantics engine. (L) compares simulation throughput for a heat transfer environment with and without the semantics engine. For 2048 and 4096 environments, the GPU overhead with the semantics engine results in FPS (Frames Per Second) reductions of only 2.97\% and 3.42\%, respectively, compared to the physics-only engine.}
\label{fig:fig_skills}
\end{figure*}

\subsection{Skill Library Results: Manipulation \text{and} Perception}
\ourmodel automates chemical experiments by providing a library of reusable and goal-conditioned skills.
These skills are generated through a combination of manipulation and perception methods.
A hierarchical state machine organizes these skills, with leaf nodes representing primitive actions and higher-level nodes representing skills composed of these primitives.
Similarly, workflows are constructed by hierarchically combining these skills.
This layered approach enables multiple levels of abstraction, facilitating collaboration between chemists and roboticists.
Roboticists can focus on developing and refining the library of abstract skills, while chemists can leverage these skills to design and implement their specific workflows.

\revise{}{\autoref{fig:fig_skills}.A illustrates a three-layer hierarchical state machine used to pick up an object and place it on top of another. This state machine is generic and can be reused by a chemist to perform pick-and-place operations across different workflow stages. The workflow proceeds from left to right and is considered complete when all actions in the state machine are successfully executed. In the diagram, dark gray indicates completed actions, light gray shows ongoing actions, and white denotes actions not yet executed.}

\paragraph{Manipulation Skills}
\ourmodel includes various motion planning and learning methods as primitive actions, serving as the fundamental building blocks for enabling manipulation skills. These methods include:
1) a robot joint-level controller,
2) a robot motion planner with an inverse kinematics (IK) solver,
3) grasping and ungrasping,
4) a whole-body controller for mobile robots,
5) relative motion in Cartesian space,
6) a waiting primitive action,
7) the cuRobo constrained motion planner with obstacle avoidance~\cite{sundaralingam2023curobo}, and
8) a \textit{check-point} primitive action for deploying trained policies (via reinforcement learning or imitation learning) on robots or automation devices.
These primitive actions can be optionally combined within a hierarchical state machine to achieve complex manipulation skills.

\autoref{fig:fig_skills} demonstrates a mobile `robot chemist' performing various manipulation skills in conjunction with other robots.
\autoref{fig:fig_skills}. A illustrates the robot picking up a well plate using the IK robot motion planner and placing it inside the OT-2 liquid handler. This example showcases how different primitive methods can be combined within a hierarchical state machine to achieve the abstract manipulation skill \textit{Pick\&Place(obj1, obj2)}, which involves picking up \textit{obj1} and placing it in \textit{obj2}. When defining the workflow, a chemist can specify the actual objects.
\autoref{fig:fig_skills}.B shows the Franka robot picking up a vial and inserting it into a vial rack hole.
\autoref{fig:fig_skills}.C presents an instance of the cuRobo motion planner reaching for a beaker inside a fume hood.
\autoref{fig:fig_skills}.D demonstrates the KUKA KMR iiwa mobile robot transferring samples between two workstations using a whole-body motion controller.
Figures~\ref{fig:fig_skills}.E–H illustrate robots performing various tasks, including pouring liquid into a beaker, weighing and scooping powders, and handing over a beaker using two robotic arms.

\paragraph{Perception}

Moving beyond traditional lab automation setups that rely on hard-coded instructions and the use of AprilTags or QR codes, \ourmodel integrates FoundationPose~\cite{wen2024foundationpose}, a foundation model for model-based 6D object pose estimation and tracking. By leveraging FoundationPose, \ourmodel reduces the need for tedious manual calibration typically required in chemistry labs to integrate multiple devices and robotic systems.
When used as the digital twin counterpart of a real-world setup, FoundationPose allows \ourmodel to determine the object's position and orientation within the digital twin configuration. This removes the need for users to manually specify object poses within the robot's workspace (as illustrated in \autoref{fig:fig_skills}.I). Furthermore, when users define an action—such as picking up an object—the estimated object pose from the perception module can be used to determine the grasping position.
Given that all assets within the digital twin possess their respective USD 3D models, these models can be used as input for FoundationPose.

\subsection{Integrated Physics \text{and} Semantics Engine}

In \ourmodel, we developed a semantics engine to simulate behaviors that are challenging to model or currently unsupported by Isaac Sim's physics engine.  
The semantics engine extends the $n$-dimensional physics state vector $\bm{x}(t) \in \mathbb{R}^n$ by introducing a set of logical semantic states $\bm{l}(t) \in \{0,1\}^m$ and continuous semantic states $\bm{s}(t) \in \mathbb{R}^p$.  
Events and processes are linked to changes in both instantaneous and continuous semantic states. These changes are functions of the physics and semantic states, as well as the input action vector $\bm{a}(t)$ from the workflow. 
The input action vector includes both continuous physics input actions (\update{\textit{e.g.,} }{}robot joint efforts) and discrete semantic actions (\update{\textit{e.g.,} }{}turning on a heater and setting its target temperature).
Both the continuous physics actions and the discrete semantic actions are updated directly on the GPU, enabling high-throughput control and simulation.
Events can occur either as the physics and semantics engines advance the simulation (\update{\textit{e.g.,} }{}two objects \update{come}{coming} into contact) or in response to an input action from the workflow (\update{\textit{e.g.,} }{}activating the heater).  
Processes represent continuous behaviours that typically require certain conditions to be met to take effect. 
For example, to aspirate a liquid into a liquid handler tip, the pipette tip must be inside the solution. Similarly, increasing the temperature of a heater requires that the heater be turned on.  
Moreover, semantic states can influence physics states. For instance, when a pipette tip is loaded, it will move in sync with the liquid handler's pipette motion.  
Together, the physics and semantics states are simulated using
$\bm{y}(t+1)= \bm{G}(\bm{y},\bm{a})$ where $\bm{y} = \begin{bmatrix}\bm{x}, \bm{s}, \bm{l}\end{bmatrix}$ is the extended physics and semantics state vector and $\bm{G}$ is the integrated physics and semantics dynamics.

\ourmodel natively supports the semantic behaviours and functionalities found commonly in chemistry labs. It simulates heat transfer and known first-order chemical kinetics using information provided by domain experts (\update{\textit{e.g.,} }{}chemists). Additionally, it simulates various device functionalities.
The semantics engine is modular and can be easily extended by users. \autoref{fig:fig_skills} (J–K) illustrate example behaviours and device functionalities enabled by the semantics engine.
\autoref{fig:fig_skills}.J shows the OT-2 liquid handler's ability to load and remove tips. When the user commands the OT-2 to load a tip, the pipette link moves toward the tip. Once the pipette is inserted into the tip, the loading occurs. After dispensing, when the user commands the OT-2 to remove the tip, the pipette moves to the OT-2’s disposal area and detaches the tip.
\autoref{fig:fig_skills}.K shows the results of aspirating and dispensing solutions during the OT-2's liquid handling process.
Moreover,~\autoref{fig:lab_environments}.F demonstrates a mobile robot turning a knob, causing water to flow from the faucet in the washing station.

The semantics engine is implemented using PyTorch, which enables GPU parallelisation.
\autoref{fig:fig_skills}.L demonstrates the computational performance of \ourmodel when the heat transfer semantics engine is added to an environment consisting of a Franka Emika robot arm, an IKA RET control heater, and a beaker.
The simulations ran on a system with an Intel Xeon W-2235 CPU (3.80GHz, 12 cores), 31Gi RAM, an NVIDIA RTX A6000 GPU, and Ubuntu 20.04.6 LTS (x86\_64).
To compare the computational performance with and without the semantics engine, the following parameters are set randomly: the robot's movements and the IKA heater's target temperature (ranging between $40-100\text{\textdegree C}$).
The results show that for 2048 and 4096 parallel environments, after running 1000 simulation steps, the GPU overhead rises by only 2.97\% and 3.42\%, respectively, when the heat transfer semantics engine is enabled.

% \begin{figure*}[!t]
%     \centering
%     \maketitle
%     \includegraphics[width=0.84\textwidth]{Images/Fig5_chemistry_experiments.pdf}
%     \captionof{figure}{}
% \label{fig:workflow_verification}
% \end{figure*}

\begin{figure*}[!t]
    \centering
    \maketitle
    \maybeincludegraphics[width=0.84\textwidth]{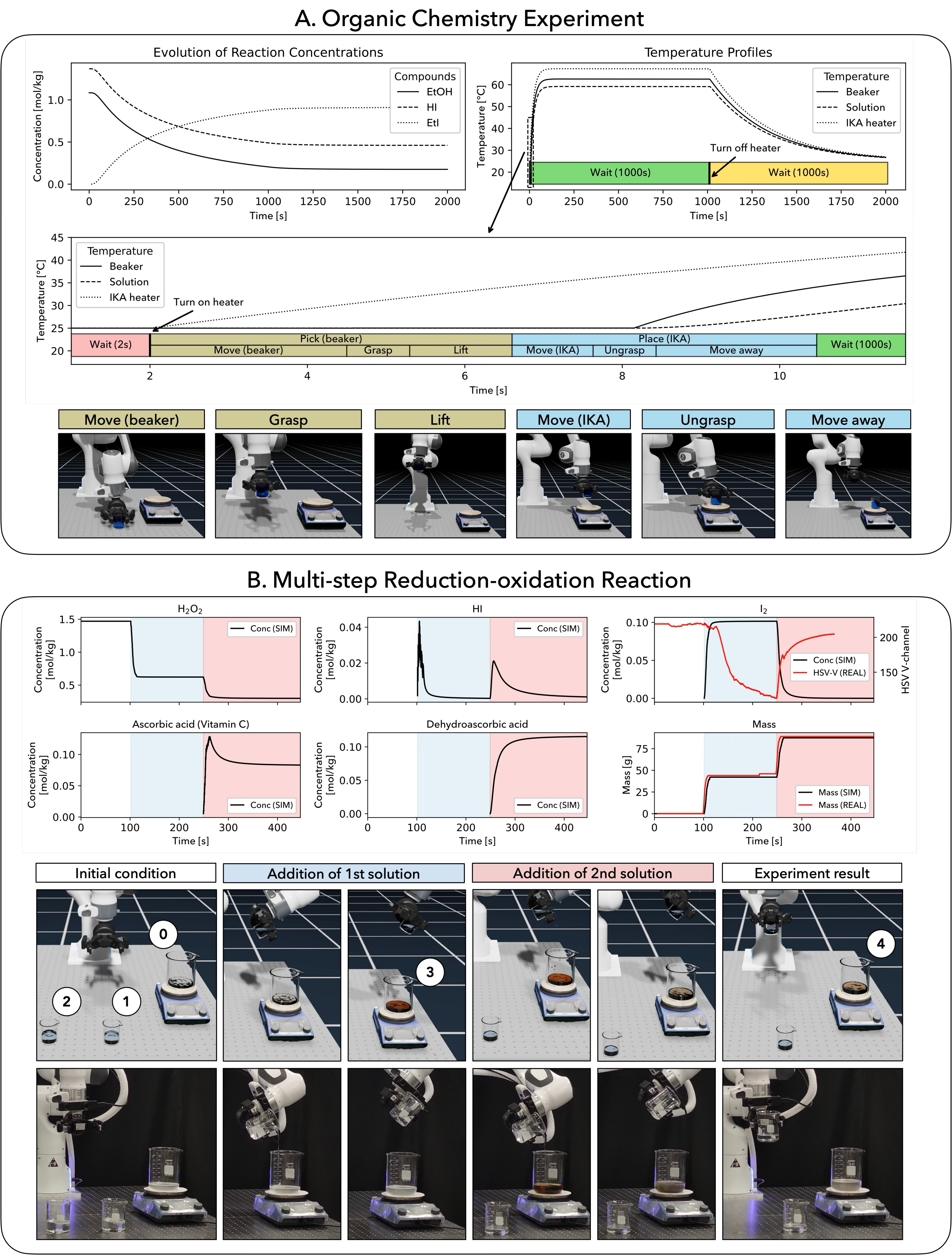}
    \captionof{figure}{\textbf{Multi-scale simulation of chemistry experiments.}
(A) A single-step organic chemistry experiment demonstrating the interaction of physical manipulation, heat transfer, and chemical kinetics simulation. (B) A two-step reduction-oxidation chemistry experiment illustrating the interaction of physical manipulation and chemical kinetics simulation. \revise{}{See \supsecref{sec:appendix:Deployment-Digital-Twin-Workflows-Real-Setup} for the video of experiment (B).}}
\label{fig:workflow_verification}
\end{figure*}

\subsection{Simulation \text{of} Multi-Scale Chemistry Workflows}
One primary purpose of digital twins is to develop and evaluate environments and workflows for chemistry experiments.
\autoref{fig:workflow_verification} shows the results of a multi-scale simulation for chemistry experiments, combining multi-body dynamics and particle simulation via the physics engine with heat transfer and chemical kinetics via the semantics engine.
\revise{}{
\supsecref{sec:appendix:chemistry-experiment-workflows} outlines the experimental workflows required to execute the procedure defined by the hierarchical state machine.
}

\autoref{fig:workflow_verification}.A illustrates an organic chemistry experiment where the unimolecular nucleophilic substitution ($\mathrm{S_N1}$) reaction occurs between ethanol ($\mathrm{EtOH}$) and hydroiodic acid ($\mathrm{HI}$), heated on the IKA heater.
\revise{The reaction only proceeds if the solution's temperature remains within the user-specified range (in this example, \([40, 70]\,^\circ\text{C}\)).
In the workflow, the heater turns on after a 2-second delay, causing only the heater's temperature to rise. At around 7 seconds, the robot places the beaker with the solution on the heater's hot plate, causing the beaker and solution temperatures to increase. Energy dissipation to ambient air affects all objects involved in the heat transfer simulation. As the solution warms, the reaction begins at 17.8 seconds and continues until 1281.6 seconds, as long as the temperature stays within the reaction's permissible range. The heater turns off at 1009.4 seconds, and the solution gradually cools due to ambient air dissipation. After 2000 seconds, the experiment concludes with a solution containing $\mathrm{EtI}$ at a concentration of 1.08~$\mathrm{mol}/\mathrm{kg}$ and a temperature of 26.6$^\circ\text{C}$.}
{
The reaction only proceeds if the preconditions are met.
This includes the requirement for the concentrations of the reactants to be non-zero, and the temperature of the solution to be within a user-specified range (in this example, \([30, 70]\,^\circ\text{C}\)).
The reaction mechanics are governed by user specified parameters, which can specify a constant $k$ reaction rate, or the parameters of a temperature-dependent $k(T)$ based on the Arrhenious equation \cite{Arrhenius1889Uber}. The temperature range precondition is still enforced to constrain the reaction to physically relevant temperatures.
In \supfigref{fig:organic_workflow_verification_40_degree},
we simulate the same process with a heater temperature of 40$^\circ\text{C}$. At this much lower temperature, the reaction rate of the substitution reaction is slower, and less product is produced by the end of the simulation.}

\revise{}{In the workflow, the heater turns on after a 2-second delay, causing only the heater's temperature to rise. At around 8 seconds, the robot places the beaker with the solution on the heater's hot plate, causing the beaker and solution temperatures to increase. Energy dissipation to ambient air affects all objects involved in the heat transfer simulation.
As the solution warms, the reaction begins at 11.4 seconds and continues until 1639.9 seconds, as long as the temperature stays within the reaction's permissible range. The heater turns off at 1010.6 seconds, and the solution gradually cools due to ambient air dissipation. After 2000 seconds, the experiment concludes with a solution containing $\mathrm{EtI}$ at a concentration of 0.91~$\mathrm{mol}/\mathrm{kg}$ and a temperature of 26.65$^\circ\text{C}$.}

\autoref{fig:workflow_verification}.B 
illustrates a multi-step redox experiment sequence, demonstrating the capabilities of the chemical kinetics simulation in handling workflows involving multiple solutions and reactions that involve both inorganic and organic transformations.
The reaction occurs at ambient temperatures, and three solutions were prepared in separate beakers.
\revise{}{The reaction starts by adding hydrogen peroxide (2.1 mmol $\mathrm{H_2O_2}$ in 48 g $\mathrm{H_2O}$ from beaker 1) with iodide (4 mmol $\mathrm{I^-}$ in 100 g $\mathrm{H_2O}$ in beaker 0). This addition oxidizes iodide, leading to the formation of iodine ($\mathrm{I_2}$), which forms dark blue-brown complex with the starch indicator over the course of 250 s. In the subsequent step,  ascorbic acid (2.4 mmol ascorbic acid in 48 g $\mathrm{H_2O}$ from beaker 2) is added to reduce $\mathrm{I_2}$ to $\mathrm{I^-}$, leading to the consumption of the iodine-starch complex and the dark blue-brown color of the reaction mixture fades away. \update{(See Supplementary Section D for experiment details)}{}
\autoref{fig:workflow_verification}.B (bottom) shows the real experiment alongside the simulated experiment. The real experiment results are displayed in the plots with red lines, representing mass changes measured by the IKA RET control scale. We also tracked the reaction progress using a camera stream focused on the solution; as the first reaction proceeds, the brightness in the HSV channel changes. As shown in the results, the first reaction occurs more quickly in the simulation than in the real experiment, highlighting the importance of online learning and parameter adaptation in the digital twin based on real experimental data.
}

\subsection{Deployment \text{to} Real Laboratory Environments}
\begin{figure*}[!t]
    \centering
    \maketitle
    \maybeincludegraphics[width=0.91\textwidth, trim=0mm 20mm 0mm 20mm, clip]{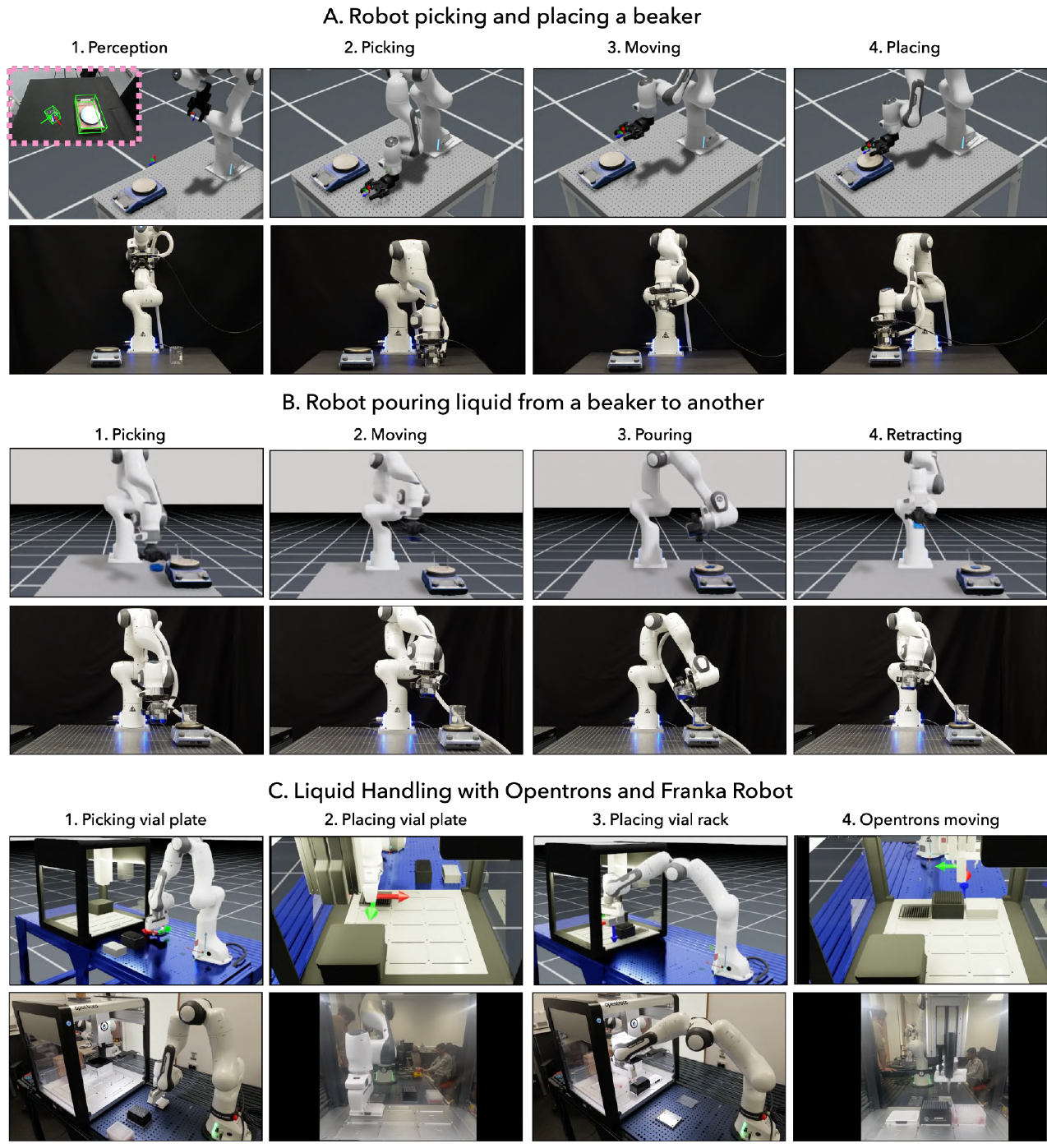}
    \captionof{figure}{\textbf{Deployment of workflows to the real setup.} These experiments demonstrate the sim-to-real transfer of workflows developed in the digital twin. (A) The robustness of pose estimation and a pick-and-place task, repeated 10 times as a common lab automation task. (B) The robot chemist transferring liquids during a pouring task. (C) A real-world chemistry lab with the robot chemist and a liquid handler working together seamlessly. Refer to \supsecref{sec:appendix:Deployment-Digital-Twin-Workflows-Real-Setup} for videos of the experiments.
}
\label{fig:task_deployment}
\end{figure*}

We evaluated \ourmodel's ability to deploy long-horizon experiments from simulation to a real-world laboratory setting.  
As shown in \autoref{fig:task_deployment}, we conducted three experiments: 1) Object Pick and Place: We assessed \ourmodel's robustness in object pick-and-place tasks using perception—a common manipulation skill in chemistry labs; 2) Liquid Handling: We demonstrated \ourmodel's ability to transfer robot manipulation skills for handling liquids; 3) Long-Horizon Experiment: We tested \ourmodel in a real chemistry lab sample preparation setup, where the robot sets up a liquid handling station with an OT-2 automation device.

\paragraph{Object Pick and Place Experiment}
A key task for robots in chemistry labs is picking and placing various objects and lab consumables, particularly transparent glassware.
\autoref{fig:task_deployment}.A shows \ourmodel performing a pick-and-place task with a beaker arbitrarily placed on a IKA plate in front of the robot as part of an experiment.
When the digital twin starts, the simulated robot syncs with the real robot. The robot observes the real scene to estimate object locations using FoundationPose and updates its digital twin. As the robot moves in simulation, the real robot follows these movements.
For this experiment, we used a state machine with IK motion planner skills.
Across 12 trials, the task execution achieved a success rate of 75\% (9 successful attempts). Failures occurred in three cases: twice due to incorrect beaker picking and once due to misplacement of the beaker on the IKA plate. These failures stemmed from multiple factors, including inaccuracies in object pose estimation, robot action errors, \revise{}{errors in camera calibration, object frame definitions,} and mismatches between digital and real assets\update{ (\textit{e.g.}, size discrepancies)}{, such as size discrepancies}.
Developing robust pick-and-place algorithms and reducing the sim-to-real gap remain active research directions. \revise{}{Addressing both human-in-the-loop sources of error and robot-level uncertainties is critical for improving autonomous reliability in real-world deployments.}

\paragraph{Liquid Pouring Experiment}
\autoref{fig:task_deployment}.B presents the robot's execution of a fundamental laboratory task: pouring fluids into laboratory glassware.
Here, the robot is pouring water into a beaker placed on an IKA RET control device. 
Initially, the robot grasps a water-filled beaker and repositions it for pouring. Subsequently, a controlled rotational motion is executed to dispense the fluid into a target beaker positioned on an IKA RET control device. The robot then returns to its home configuration, completing the fluid pouring task.
\revise{}{
A hierarchical state machine was used to run the experiment with an inverse kinematics controller. In total, we conducted 10 experiments, with the beaker placed at arbitrary locations within the workspace. The robot estimated the beaker's position using FoundationPose. We achieved a 90\% success rate in real-world experiments. One failure occurred due to a large error in the beaker pose estimation, which caused the robot to grasp the beaker incorrectly, resulting in the beaker sliding. Additionally, there was a case where the simulation failed but the real experiment succeeded: the beaker fell off the table in the simulation because the simulated table size did not match the real one. This highlights the importance of accurately matching the simulated environment to the real robot workspace.
The amount poured in the real experiment, measured using the IKA plate, was $59.6 \pm 7.9$\,mL out of an initial volume of 101.77\,mL, while in the simulation it was $61.3 \pm 4.4$\,mL out of 100\,mL. The difference in pouring amounts between the simulation and the real experiment stems from differences in the initial liquid volumes, slight variations in beaker sizes, and the differing behavior of the liquid in the physical world versus the simulation.
}

\paragraph{Liquid Handling Station}
This experiment demonstrates \ourmodel's capability in robotics-assisted liquid dispensing.
\autoref{fig:task_deployment}.C shows the Franka Research 3 (FR3) robot setting up the station by placing the dispensing rack, tips rack, and reservoir well in the OT-2 tray. The OT-2 head then attaches to the pipettes before moving to the dispense rack to start fluid dispensing.

\ourmodel programs the task in simulation, and the policy is transferred zero-shot to the OT-2 and FR3. \revise{}{The policy was implemented as a hierarchical state machine with hand-tuned low-level controllers and inverse kinematics, developed and refined entirely in simulation to ensure robust deployment on the real system.} 
The real FR3 and OT-2 are actuated using \texttt{panda-py}~\cite{Elsner2023taming} and \texttt{ot2\_driver}~\cite{Vescovi2023}, respectively.

% \begin{figure*}[!h]
%     \centering
%     \maketitle
%     \includegraphics[width=0.91\textwidth, trim=0mm 20mm 0mm 20mm, clip]{Images/Fig6_sim2real.pdf}
%     \captionof{figure}{}
% \label{fig:task_deployment}
% \end{figure*}

 \section{Discussion}\label{sec:discussion}

This work introduces a multi-scale digital twin simulation framework\update{}{for chemistry labs.} \update{that combines high-fidelity, photorealistic rendering with efficient robotic design to accelerate the development of autonomous chemistry laboratories.
The framework simulates long-horizon chemistry experiments, incorporating robot manipulation, fluid and granular material simulation via a physics engine, as well as heat transfer, chemical kinetics, and device functionalities through a semantics engine.
Its modular architecture enables the creation of multi-stage, complex laboratory digital twins, while supporting prominent robot learning paradigms such as reinforcement learning, imitation learning, and motion planning. The simulator integrates seamlessly with curated laboratory assets, commonly used robots, and automated platforms in autonomous labs.}{} While there are alternative robotic simulation versions for chemistry lab tasks, most either focus on a single task~\cite{Kadokawa2023PowderWeighing, Pizzuto2024Scraping, LopezGuevara2020StirToPour} and the most comprehensive one to date lacks heterogeneous robots with automated platforms, and real-world sim-to-real transfer~\cite{Li2024chemistry3droboticinteractionbenchmark}. 
In comparison,~\ourmodel distinguishes itself through its integrated semantics engine, hierarchical workflow representation and easy sim-to-real transfer of developed systems. Additionally, it simplifies usability by offering an extensive library of assets, manipulation skills, and perception tools for defining digital twin environments and generating workflows for chemistry experiments.

However, the current iteration is not without limitations. 
The reliance on approximated chemical reaction models within the semantics engine can introduce inaccuracies. 
Integrating quantum chemistry computational methods and machine learning could strike a balance between accuracy and the efficiency required for real-time simulation~\cite{johnson2024machine}.
\revise{}{Currently, \ourmodel does not support gas-phase simulation in chemical reactions; future versions aim to incorporate this using Isaac Sim's physics engine and semantics integration.}
Additionally, the design and computational cost of simulating long-horizon experiments remains a challenge, especially for successfully planning extended workflows.
Incorporating large language and vision models could lower the barrier to entry, increasing the accessibility of \ourmodel for designing digital twin environments and automating workflow planning~\cite{black2024pi0VLA}. Regarding workflow planning, while hierarchical state machines offer an intuitive and user-friendly way to define workflows, they have limitations in deployment—particularly in recovering from failures or generating contingency plans. To address these challenges, more flexible frameworks such as behavior trees~\cite{colledanchise2018behavior} or and-or graphs~\cite{darvish2020hierarchical} could be adopted.
Furthermore, the reliance on FoundationPose may introduce inaccuracies in object pose estimation, affecting real-world deployment precision.
Refinement of object pose estimation techniques, potentially through the integration of multimodal sensor data, will improve real-to-sim deployment performance.

\revise{}{In addition to improving workflow flexibility and perception accuracy, addressing real-world experimental failures, such as clogging or equipment malfunction, is critical for robust and safe deployment. These failures often arise from discrepancies between simulated and physical environments, caused by differences in states, sensor noise, model inaccuracies, or limitations of the physics and semantics engines. Future versions of \ourmodel will integrate real-time feedback from IoT sensors, such as temperature probes, flow meters, and visual monitoring, to detect deviations between simulation and reality. These measurements can trigger safety protocols, flag simulation inaccuracies, and guide model refinement~\cite{leong2024steering}. By incorporating online learning methods, surrogate models trained from real-world data can be embedded into the digital twin to improve fidelity and enable predictive maintenance strategies that proactively address failure scenarios~\cite{liu2023reflect}.}

Future work could expand the skill library to include manipulation policies for soft materials\update{ (e.g., tubing)}{, such as tubing,} and delicate glassware. Additionally, large language models (LLMs) could utilize the skill library to procedurally generate and execute hierarchical state machine plans in simulation, forming a data-generation pipeline for training experimental planner policies~\cite{yu2023language, skreta2024replan, wang2023robogen}. Beyond this, this framework could facilitate the co-design of chemistry workstations and laboratories to optimise performance metrics such as throughput while also serving as a tool for data collection to train experimental planning models.

\revise{}{Beyond automation, digital twins could serve as platforms for scientific exploration and discovery. By enabling parallel simulations and ``what-if'' experimentation, they support the development of AI-driven experimental planners and hypothesis testing in silico. In domains where physical processes, such as fluid handling or reaction kinetics, can be modelled with reasonable accuracy, digital twins allow exploration of \update{novel}{new} protocols and experimental conditions. Despite current limitations in simulating complex chemistry or multiphase interactions, these platforms offer a foundation for training robust robotic policies that can later be deployed for real-world scientific discovery and experimentation.}

We believe~\ourmodel is a powerful tool for accelerating chemical discovery and foresee its application in adjacent fields, such as biotechnology. It represents a \update{significant}{substantial} effort to bridge the robotics and lab automation communities by offering a platform that could be used to create benchmarks for testing robotic approaches in the lab automation application domain. Additionally, it provides a tool for the lab automation community to leverage these approaches, enabling the development of flexible, robust, and safe algorithms, as well as data collection, testing, and monitoring for chemistry lab automation and scientific discovery. Future developments will further solidify its role as a key resource for researchers and industry stakeholders, accelerating the deployment of automated robotic chemistry experiments. Towards this goal, we encourage open-source contributions to expand its capabilities and foster collaboration.

\section{Methods}\label{sec:methods}

The components and architecture of \ourmodel are presented in \autoref{fig:architecture}. \ourmodel constructs the simulation environment using user input and an asset library, then simulates multi-scale chemistry experiments in the digital twin while verifying the workflow.
For real-world deployment, \ourmodel captures environmental images and robot proprioception data to execute policies across laboratory devices and robotic systems.

The simulation framework is formulated on the assumption that a set of agents\update{ (\textit{e.g.,} a robotic chemist)}{, in our case robotic chemists,} interact with the world\update{, which is the chemistry lab environment, receives observations from its environment and interacts with its environment through computed actions.}{—the chemistry lab environment—receiving observations and acting through computed controls.}
The world comprises robotic chemists, which could be the fixed arms, the mobile manipulator or even an automation platform, its sensory elements and surrounding objects, including laboratory instruments, glassware and particles.
The world description is inherently modular and can be set through standardized scripts or Universal Scene Description (USD) using Nvidia Isaac Sim's graphical interface.

This section delineates the methods of the four main components of \ourmodel, 1) digital twin environment generation and initialization, 2) multi-scale simulation with the integrated physics and semantics engine, 3) workflow plan generation with manipulation skill library, perception, and hierarchical state machine, 4) deployment of the simulation to the real setup.

\subsection{Digital Twin Environment Generation}
One of the main bottlenecks in the usage of digital twins for chemistry lab automation is the effort required in the generation of those virtual environments and the skill gap between the chemistry domain and digital twin development requirements. To lower this barrier, \ourmodel comes with different methods, including: 1) providing methods to generate digital twin assets and a rich library of assets with commonly used objects and tools in chemistry labs, and 2)  methods to generate digital twin environments with ease.

\paragraph{Asset Generation}
\ourmodel comes with a library of assets, ready to be used for lab automation purposes and to learn new policies in simulation. The following asset types are supported: 
glassware, laboratory tools and devices, fixed-based and mobile robots, automation platforms, and lab spaces.
Besides rigid and articulated assets, it supports liquid and powders as particle systems, and soft objects.

Digital twin assets are represented using Universal Scene Description (USD)\cite{openusd_software_2025}, a high-performance platform initially developed for film and visual effects. USD supports robust data interchange via comprehensive schemas for geometry, shading, lighting, and physics, and its unique composition features enable flexible asset assembly and collaborative workflows. Assets in \ourmodel incorporate physical and textual properties, and users can attach custom attributes—such as object key points and coordinate frames—to enhance simulation and automation tasks.

Physical assets in our digital twin simulation include prims (collision-only static objects), rigid bodies, articulated bodies, robots, particle systems (fluids and powders), and soft objects. In Isaac Sim, powders and fluids are modeled using a position-based dynamics (PBD) solver~\cite{macklin2013position} integrated with an iterative density solver that enforces constant density via positional constraints. This approach enables large time steps suitable for real-time applications. Consequently, PBD provides a flexible platform for simulating phenomena such as liquids, granular materials, cloth, and deformable bodies.

\ourmodel leverages multiple methods to generate the asset library. One approach is to convert 3D asset models—either user-designed or sourced online—into USD assets, with tools like Isaac Sim and Blender available for further modification. Robots and articulated models, often described using URDF (Unified Robot Description Format), can be directly converted to USD in Isaac Sim.  
Additionally, \ourmodel utilizes advances in computer graphics for 3D scene reconstruction. We employ Neural Radiance Field (NeRF)~\cite{mildenhall2021nerf} to generate 3D representations of rigid assets by optimizing a continuous volumetric scene function from sparse input views (\update{e.g., }{}images or video scans). Nerfstudio~\cite{tancik2023nerfstudio} simplifies this process for users with its modular framework and intuitive API. Online platforms like Luma AI~\cite{luma_ai_2025_online} also offer tools for creating 3D assets.
Users manually define core physical properties, including mass. For additional parameters\update{ (\textit{e.g.,} friction coefficient)}{, such as friction coefficient}, default values and materials can be refined for any asset—whether NeRF-generated or sourced from 3D assets.

\paragraph{Environment Generation}
A chemistry digital twin may consist of numerous assets, making environment generation challenging. To simplify this, \ourmodel provides an asset library where users can select reusable models for common lab items such as beakers, heaters, and robotic arms. As shown in~\autoref{fig:architecture} and~\autoref{fig:lab_environments}, users can add assets by selecting them from the library and specifying their location in the environment.
The USD scene of an asset in \ourmodel may include multiple objects, such as a rack with pipette tips, reducing duplication and simplifying the creation of complex setups like the full lab environment in \autoref{fig:lab_environments}. Users can also update additional details such as keypoints or physical properties, ensuring accurate simulation behaviour.

Digital twin environments are created for two main purposes: synthetic environment generation and real-to-sim replication. The synthetic environment is useful for the feasibility analysis of a workstation before building it in the real world or for developing and testing new methods, such as experimental planning algorithms. 
A real-to-sim environment replicates a real lab setup, supporting tasks like monitoring and training machine learning models for automation.
\ourmodel provides interfaces to connect digital twin environments with real lab sensor data\update{ (\textit{e.g.,} camera)}{, such as camera input,} and to control robots, as explained later in this section. Additionally, \ourmodel supports heterogeneous lab setups inherently by enabling the definition of multiple robotic systems (agents), including robots, automation platforms, and other devices, allowing concurrent task execution in simulated environments.
This multi-agent capability facilitates research in scheduling and experimental planning for lab automation, paving the way for high-throughput solutions to chemistry automation tasks.

\subsection{Multi-Scale simulation: Integrated Physics  \text{and} Semantics  Engine}
\update{Many real-world physical and chemical dynamics are challenging to simulate, such as the locking mechanism of a door or the aspiration and dispensing of liquids by a liquid handler, and are often unsupported by robotics simulators (\textit{e.g.,} heat transfer and chemical reactions).}{Many real-world physical and chemical dynamics are challenging to simulate, ranging from the locking mechanism of a door and liquid handling tasks to processes often unsupported by robotics simulators, such as heat transfer and chemical reactions.}
Despite the ubiquity of such tasks in daily life and chemistry lab activities, existing robotic simulation tools do not support them.  
Furthermore, some assets in the digital twin possess specific functionalities. For instance, a heater, when \texttt{?heaterOn $\leftarrow$ true}, acts as a heat source. By integrating these device functionalities, digital twins can effectively manage long-horizon experimental plans.
The semantics engine extends the Isaac Sim physics engine by enabling the simulation of these behaviors using approximated models. These behaviors can influence the physics engine's simulation state, and conversely, the physics state can impact the semantics engine. For instance, the amount of water flowing from a faucet in a chemistry lab washing station is linked to the degree to which the faucet handle is rotated (\autoref{fig:lab_environments}.F).

The semantics engine can also serve as a source of sparse rewards for training reinforcement learning (RL) policies, aiding in the exploration of long-horizon tasks. Implemented in PyTorch with GPU parallelization, the semantics engine enables efficient execution of RL algorithms. It is designed to be modular, allowing users to add custom behaviours. \ourmodel offers a semantic behaviour library for common chemistry workflows (refer to \autoref{sec:results}), including liquid handling, heat transfer, and chemical reaction approximations.

The semantics engine is modeled with the logical state vector $\bm{l}(t) \in \{0,1\}^m$ and continuous states $\bm{s}(t) \in \mathbb{R}^p$, which extends the physics engine's state vector $\bm{x}(t) \in \mathbb{R}^n$. The input action vector $\bm{a}(t)$ to the \ourmodel simulation consists of both continuous and logical elements, such as robot joint angles for picking up an object or turning on a heater and setting it to a target temperature.  
Events $\bm{l}(t+1) = \bm{E}(\bm{x}, \bm{s}, \bm{l}, \bm{a})$ and processes $\bm{s}(t+1) = \bm{P}(\bm{x}, \bm{s}, \bm{l}, \bm{a})$ are linked to changes in both instantaneous and continuous semantic states, thereby extending the physics simulation $\bm{x}(t+1) = \bm{F}(\bm{x}, \bm{a})$. The logical states $\bm{l}(t)$ can influence the behavior of physics dynamics $\bm{F}$ and semantic processes $\bm{P}$, evolving over time as the logical states change: $\bm{F}_t$ and $\bm{P}_t$. This can be achieved by either directly updating the models $\bm{F}$ and $\bm{P}$ or setting preconditions for $\bm{P}$ to take effect.  
For example, loading a pipette tip into the liquid handler requires real-time changes to $\bm{F}$, so that when the pipette tip is loaded, it moves in sync with the liquid handler's pipette motion. Similarly, aspirating liquid into the pipette tip requires the tip to be inside the solution. Increasing a heater's temperature, on the other hand, requires the heater to be turned on.

\ref{alg:semantics-physics} in ~\supsecref{sec:physics-semantics-engine-algorithm} outlines the integration of the Physics and Semantics engines.  
% At each simulation step, we first pass the current physical states $\bm{x}_{t}$ and actions $\bm{a}_{t}$ to the physics engine to compute the next physics state. Next, we iterate over all the semantics processes and events to estimate the next semantic states $\bm{s}_{t+1}$ and $\bm{l}_{t+1}$. Additionally, events can influence the behaviours of both the physics and semantics.  
% \begin{algorithm}[H]
% \scriptsize
% \begin{algorithmic}[1]
% \caption{\textsc{Physics\&SemanticsEngine()}}
% \label{alg:semantics-physics}
% \Require {$\bm{x}_{0},\bm{s}_{0},\bm{l}_{0}, \bm{a}_{t} $ }
% \Ensure {$\bm{x}_{t+1},\bm{s}_{t+1},\bm{l}_{t+1}$}
% \ForAll{$t \in \mathcal{T}$} 
% \State $\bm{x}_{t+1} \leftarrow$  $\bm{F}_{t}$($\bm{x}_{t}, \bm{a}_{t}$) \Comment{physics engine}
% \State $\bm{s}_{t+1} \leftarrow$  $\bm{P}_{t}$($\bm{x}_{t+1}, \bm{s}_{t}, \bm{l}_{t}, \bm{a}_{t}$) \Comment{semantic processes}
% \State $\bm{l}_{t+1}, \bm{F}_{t+1}, \bm{P}_{t+1} \leftarrow$  $\bm{E}$($\bm{x}_{t+1}, \bm{s}_{t+1}, \bm{l}_{t}, \bm{a}_{t}$) \Comment{semantic events}
% \EndFor
% \end{algorithmic}
% \end{algorithm}
While this formalization facilitates the scaling of new behaviours by the semantics engine, \ourmodel also allows users to customize interactions between the physics and semantics engines.

Given that the simulation of heat transfer and chemical kinetics is crucial for simulating complete chemistry workflows, their descriptions are provided in the following paragraphs.

\paragraph{Heat Transfer Simulation}
\ourmodel employs an energy balance approach to predict temperature variations in different components over time, considering conduction and convection as the primary heat transfer modes. Thermal interactions with the surrounding air are modeled through convection, while heat transfer between contacting objects is handled via conduction. The conductive heat transfer between two contacting objects and the heat transfer through convection are modelled by:
\begin{equation}
\dot{Q}_{cond} = \frac{k~A}{d} \Delta{T} \quad \text{and} \quad 
\dot{Q}_{conv} = h~A \Delta{T}~,
\end{equation}
where $T(K)$ represents the temperature, $k (W/m \cdot K)$ is the thermal conductivity, $A (m^2)$  denotes the heat transfer area, $d (m)$  is the thickness of the conductive medium, $h (W/m^2 \cdot K)$  is the convective heat transfer coefficient, \(\dot{Q}\)($W$)  represents the heat transfer rate.
Additionally, heat generated by a heat source \update{(e.g., a heater) }{}is modelled by:
\begin{equation}
    \dot{Q}_{gen} = K_{gen} ({T_{target}- T(t)})~,
\end{equation}
where $K_{gen} (W K^{-1})$ is the heat source constant, and  $T_{target} (K)$ is the target temperature of the heater.
By applying the first law of thermodynamics, we derive the governing equations that describe the rate of temperature change for different objects in the digital twin as follows:
\begin{equation}
    m C \frac{dT }{dt} = \Sigma{\dot{Q}}~.
    \label{eq:1st-law-thermo}
\end{equation}

During the simulation, the heat transfer rate values on the right-hand side of \autoref{eq:1st-law-thermo} are determined by the semantic logical states, such as the contact between different objects and whether the heater is turned on. Further details on the heat transfer simulation, along with assumptions and an example, can be found in \supsecref{sec:appendix:heat-transfer-semantics-engine}.

\paragraph{Chemical Kinetics Simulation}
\ourmodel provides approximate reaction kinetics simulations to estimate the change in concentrations of reactants and products as a function of time. This not only verifies the occurrence of a reaction but also allows for multi-step reactions, in which the product of an earlier reaction is the reactant for a subsequent reaction.
\revise{}{For a given reaction, the user specifies the reaction template to transform the reactants in a container into product compounds, and also approximate parameters for the kinetics simulation, including the reaction order and the parameters governing the temperature-dependent reaction rate constant, $k$.}
\revise{For a given reaction, the user specifies the reaction template to transform the reactants in a container into product compounds, and also approximate parameters for the kinetics simulation, including the reaction order and the reaction rate constant $k$. For an example reaction $A + 2B \rightarrow AB_2$, the reaction rate would be first order in the concentration of A, denoted $[A]$, and second order in $[B]$.  The reaction rate at a given time would be}{}
\revise{while the product concentration reaction rate would be the negative of the reactant concentration reaction rate,}{}
\revise{The reaction would continue until the preconditions are no longer met, such as a change in temperature, or the depletion of a reactant species.}{}

\revise{This reaction simulation has many limitations, which can be addressed in future improvements. In particular, the dependence of the reaction rate constant on temperature is neglected. The reactions that can be simulated are limited to forward-only reactions which can be represented using SMARTS templating. We reiterate that the reaction verification methods are only semi-quantitative, and more sophisticated system-specific reaction kinetics simulation and quantum chemical calculations are required for high-quality quantitative predictions.}{}

\revise{}{To incorporate temperature dependence, $k$ is calculated at each time step based on the system temperature ($T$) using the Arrhenious equation~\cite{Arrhenius1889Uber},}
\begin{equation} \label{eq:arrhenius}
\revise{}{k = A \exp\left(-\frac{E_a}{RT}\right).}
\end{equation}
\revise{}{Here, $A$ represents the pre-exponential factor, $E_a$ is the activation energy for the reaction, and $R$ is the universal gas constant. It should be noted that the parameters $A$ and $E_a$ are estimated based on prior experimental knowledge from organic chemists and should not be taken as exact quantities. To maintain numerical stability, the temperature is constrained to a physically relevant temperature range.}

\revise{}{For an example reaction $A + 2B \rightarrow AB_2$, the reaction rate would be first order in the concentration of A, denoted $[A]$, and second order in $[B]$. The reaction rate at a given time would be}
\begin{equation}
\revise{}{\dot{[A]} = \frac{1}{2} \dot{[B]} = - k [A][B]^2,}
\end{equation}
\revise{}{while the product concentration reaction rate would be the negative of the reactant concentration reaction rate,}
\begin{equation}
\revise{}{\dot{[AB_2]} = + k [A][B]^2.}
\end{equation}
\revise{}{The reaction would continue until the preconditions are no longer met, such as a change in temperature, or the depletion of a reactant species.}

\revise{}{While the inclusion of temperature effects enhances the simulation's physical fidelity, several limitations remain which can be addressed in future improvements. The reactions that can be simulated are limited to forward-only reactions which can be represented using SMARTS templating. We reiterate that the reaction verification methods are only semi-quantitative, and more sophisticated system-specific reaction kinetics simulation and quantum chemical calculations are required for high-quality quantitative predictions.}

\paragraph{\update{}{Reduction-Oxidation Execution Protocol}}
\update{}{The following stock solutions were prepared manually using an analytical balance, all chemical were purchased from Sigma Aldrich.}

\update{}{\textbf{Solution 0:} $\mathrm{KI}$ (0.66 g, 4 mmol) and saturated starch in 100 g $\mathrm{H_2O}$, total concentration of $\mathrm{KI}$ = 0.04 mol/kg}

\update{}{\textbf{Solution 1:} $\mathrm{H_2O_2}$ (3 g of 5\% w/w, 4.4 mmol) in 100 g $\mathrm{H_2O}$, total concentration of $\mathrm{H_2O_2}$ = 0.044 mol/kg}

\update{}{\textbf{Solution 2:} ascorbic acid (0.88 g, 5 mmol) in 100 g $\mathrm{H_2O}$, total concentration of ascorbic acid = 0.05 mmol/kg}

\update{}{\textbf{Step 1:}}
\update{}{To a large reaction beaker (\textbf{Beaker 0}) containing 100 g of \textbf{Solution 0} (4 mmol $\mathrm{I^-}$, saturated starch) over an analytical balance, the robotic arm picked up \textbf{Beaker 1} containing \textbf{Solution 1} and poured around 50 g \textbf{Solution 1} into the reaction beaker. An increase in total mass was measured to be 48 g, total amount of $\mathrm{H_2O_2}$ added was calculated to be 2.1 mmol.\newline}
\update{}{The mixture was left for reaction for 50 s, and the color gradually turned dark blue due to the oxidation of $\mathrm{I^-}$ into $\mathrm{I_2}$ which immediately formed a dark blue-brown iodine-starch complex.}

\update{}{\textbf{Step 2:}}
\update{}{To the reaction beaker containing product of \textbf{Step 1} , the robotic arm picked up \textbf{Beaker 1} containing \textbf{Solution 2} and poured around 50 g \textbf{Solution 2} into \textbf{Beaker 0}. An increase in total mass was measured to be 48 g, total amount of ascorbic acid added was calculated to be 2.4 mmol.}

\update{}{The mixture was left for reaction for 50 s, and the dark blue-brown color gradually faded due to the reduction of $\mathrm{I_2}$ into $\mathrm{I^-}$, which lead to the subsequent consumption of the iodine-starch complex.\newline}

\paragraph{System and Workflow Verification}
Following the semantics engine description and the workflow logic definition of a chemical workflow, we demonstrate a chemical reaction verification protocol that ensures the simulated execution aligns with expected chemical outcomes. This protocol serves as a verification for both the simulated system and the generated workflow. This protocol tracks the states of compounds within containers, maintaining both identity and concentration throughout the simulated robotic manipulations and reactions. Before executing a user-defined reactions, preconditions are checked, such as the temperature or presence of reactants. Upon successful precondition checks, reactant compounds are manipulated using cheminformatics templating to produce product molecular compounds. The simulator applies approximate estimations of concentration changes resulting from the reaction, updating the simulated container contents accordingly. It is important to emphasize that these concentration estimations are not intended for quantitative accuracy, which is beyond the scope of this workflow verification. Instead, they provide a qualitative confirmation that the expected reaction has occurred within the defined workflow, aiding in debugging and validation.

This chemical reaction workflow verification protocol serves a dual purpose in the context of automated laboratory procedures. First, it provides a valuable training tool for autonomous agents tasked with executing chemical workflows. By employing a task-based verification method, agents can receive feedback on the correctness of their actions within the simulated environment. This feedback loop is crucial for developing robust and reliable automated agents capable of navigating complex laboratory tasks. Second, the protocol offers chemists a tool for evaluating and debugging in silico representations of their workflows. Even without precise reaction kinetics, the approximate concentration changes and precondition checks provide valuable insights into the feasibility and logical flow of the designed workflow. This allows chemists to identify potential errors or inefficiencies in their experimental plans before physically implementing them in the laboratory.

\subsection{Workflow Plan Generation}
\ourmodel generates workflow plans by integrating various manipulation and perception methods within a hierarchical task definition framework to construct a skill library. This library enables the easy creation of hierarchical workflows for executing chemistry experiments.

\paragraph{Manipulation Methods and Skill Library}
For autonomous execution of chemistry experiments, robotic systems and automation platforms\update{ (\textit{e.g.,} liquid handlers)}{, such as liquid handlers,} must be capable of perceiving their environment and manipulating objects and devices. To facilitate this, we developed a skill library that allows users or AI agents to leverage predefined skills to perform long-horizon experiments with multiple sequential steps.  
The \ourmodel skill library comprises both classical motion planning methods and learning-based policies, all of which are parallelizable on GPUs. These components are described as follows.

\textit{Differential inverse kinematics (IK) and whole-body mobile robot controller.}
We employ the damped least squares method to solve the inverse kinematics problem~\cite{buss2004introduction}. This approach utilizes the Jacobian mapping from joint-space velocities to end-effector velocities to compute incremental changes in the joint space, guiding the robot's end effector toward a desired pose.  
The mobile robot is modelled as a floating base, with different controllers used to move either its base, only the upper body (arm manipulator), or the entire body. In the case of whole-body control, the Jacobians of the robot's end effector are computed relative to the inertial or world frame~\cite{RD-Lecture-Notes}.

\textit{cuRobo motion planner.}
To ensure safe robot motion in chemistry laboratories, we employ cuRobo, a collision-free motion generation framework that computes optimized robot trajectories based on the initial configuration and target end-effector pose. The system operates in two key stages: first, solving a parallelized collision-free inverse kinematics (IK) problem to determine feasible goal configurations, and second, leveraging GPU-accelerated optimization with parallelized seed trajectories to generate motion solutions. The trajectory optimization formulation incorporates multiple constraints, including joint position, velocity, acceleration, and jerk limits, as well as self-collision and environmental collision avoidance, while ensuring the target end-effector pose is reached at the final timestep.

\textit{Model-free reinforcement learning.}
\noindent For policy training, we employ Proximal Policy Optimization (PPO)~\cite{schulman2017proximal}, a policy gradient reinforcement learning method utilizing first-order optimization. PPO maintains an effective balance between data efficiency, reliable convergence properties, and implementation simplicity, making it well-suited for robotic applications requiring computational efficiency and robust performance. To train long-horizon policies, \ourmodel employs a curriculum scheduling strategy~\cite{wang2021survey} that begins with simplified short-horizon tasks and progressively increases both task complexity and temporal scope.

\paragraph{Perception}
For object pose estimation in chemistry laboratories, we employ FoundationPose~\cite{wen2024foundationpose}, a vision foundation model unifying model-based and model-free perception that requires no fine-tuning for deployment. As our digital twin simulation environments already possess CAD models of all target objects, the system integrates seamlessly. 
FoundationPose first samples initial poses uniformly around the object, refines them via a neural network, and selects the optimal pose with the highest score.

\paragraph{Hierarchical State Machine Task Representation}
Automating a chemistry experiment requires collaboration between chemists and automation researchers. Chemists define experimental workflows, while roboticists provide task descriptions and policies that enable structured execution. A chemistry experiment may span multiple workstations or labs, and tasks often consist of hierarchically organized subtasks. To accommodate this complexity, \ourmodel represents workflows using a hierarchical state machine, allowing arbitrary levels of abstraction.  

A workflow is modeled as a tree \( T = \langle N, E \rangle \), where \( N \) represents nodes (actions), and \( E \) denotes edges connecting them. The tree depth \( d \) corresponds to the level of abstraction, with parent nodes representing higher-level tasks and child nodes refining them into specific actions. The leaves of the tree are primitive actions, where execution methods and robotic manipulation policies are implemented. A workflow is completed when all nodes are executed and its goal is achieved. A node is considered executed once all its child nodes have been processed. This structured approach enables seamless integration of policies at different levels of abstraction. 
\autoref{fig:fig_skills} illustrates the construction of a reusable skill by integrating manipulation and perception methods within a hierarchical tree structure for task definition. These skills can, in turn, be utilized to create long-horizon workflows.

\subsection{Deployment to Real Setup}  
Upon successful task execution in the simulated digital twin of the chemistry environment, the solution is deployed to the real setup. During deployment, \ourmodel updates the simulated environment with the real setup by interfacing the physical robots and devices, whose states are initially updated in the digital twin to match the real setup. The robot, equipped with an end-effector-mounted camera (or alternatively, a scene camera), observes the lab environment and estimates object poses using FoundationPose, which is then used to update the digital twin. Following initialization, the workflow generates actions for the robot and devices in the digital twin, with updated states passed to their physical counterparts for execution after each simulation step.

We employ the real robot in impedance control mode to compensate for minor discrepancies between simulation and reality, particularly during contact-intensive tasks such as pick-and-place or insertion operations. \ourmodel incorporates force feedback from the robot as a safety mechanism, halting task execution if predefined safety thresholds are exceeded. The synchronized digital twin provides a powerful tool for real-time debugging, visualization, and monitoring, enabling chemists to track the experiment's progress and facilitating development. Once experimental workflows are validated in simulation, an alternative deployment strategy involves directly executing the verified workflows on the physical systems (robots and devices) without relying on the digital twin.

\section{Data availability}

All source data required to evaluate the presented conclusions are available within the paper and in the Supplementary Materials. All supporting data were generated using the \ourmodel code.
\update{}{Source data for Figures 4-6 are available with this manuscript.}

\section{Code availability}The code for \ourmodel can be found on Zenodo\cite{Darvish2025MatterixCode} and at {\small \url{https://github.com/AccelerationConsortium/Matterix}}.

\section{Acknowledgements}
\label{sec:Acknowledgements}
The authors would like to thank Adam Edwards for helping with code documentation and reproducibility, and Daniel Hatcher \update{}{and Yang Cao} for helping with asset creation.
We would like to thank Alan Yuan for his contributions to an early version of the semantic framework; Artur Kuramshin, Hyunjin Kim, and Kevin Thomas for their assistance with asset and environment generation; and Marta Skreta, Jiaru Bai\update{}{and Charlotte Boott} for their insightful discussions and feedback.
This work was supported by the University of Toronto’s Acceleration Consortium from the Canada First Research Excellence Fund, grant number CFREF-2022-00042, the Leverhulme Trust through the Leverhulme Research Centre for Functional Materials Design, the Engineering and Physical Sciences Research Council (EPSRC) under grant agreements EP/V026887/1 and EP/Y028759/1, the European Research Council (ERC) under the European Union’s Horizon 2020 research and innovation programme (grant agreement No 856405), the Royal Society via a Research Professorship (RSRP/S2/232003), and the Royal Academy of Engineering under the Research Fellowship Scheme.
 \section{Author Contributions}
\label{sec:AuthorContributions}
Authors are listed alphabetically by contribution.

\textbf{Designed and built the core infrastructure:} A.M., A.S., G.P., H.F., K.D., M.B.

\textbf{Rigid assets library:} A.M., A.S., B.Z., H.F., Je.C., Jo.C., K.D., M.B., N.R., Z.Z.

\textbf{Articulated asset library:} A.M., A.S., Jo.C., K.D.

\textbf{Robot assets:} A.M., A.S., K.D., Z.Z.

\textbf{Full lab assets:} A.M., A.S., Jo.C., K.D.

\textbf{Nested rigid assets:} Jo.C., K.D.

\textbf{Environments:} A.M., A.S., H.F., Je.C., K.D., N.R., Z.Z.

\textbf{Semantics engine:} Chemistry (A.S., G.T., H.H., K.D.), Device Functionalities (A.S., K.D.), Heat Transfer (A.S., H.D., K.D.).

\textbf{Particle system} (Fluids and Powders): A.S., K.D., M.B., N.R.

\textbf{Skill library:} FoundationPose(A.M., S.H., Y.Z.), IK solver (K.D., M.B.), RL(A.M., Je.C., K.D., M.B.), Whole body controller (K.D., Z.Z.), cuRobo (K.D., Y.W.).

\textbf{Deployment to real setups:} FR3 and OT-2 liquid handling experiment(A.W., H.F., N.R., S.V.), Liquid pouring experiment (A.M., Y.Z.), Pick-and-place experiment (A.M.), Sim-to-real deployment pipeline (A.M., H.F., K.D.).

\textbf{Writing the paper draft:} G.P., H.D., H.F., K.D., M.B., Y.Z., Z.Z.

\textbf{Supervision:} A.A.G., A.G., A.I.C., F.S., G.P., H.F., K.D., M.B.

 \section{Competing Interests}
\label{sec:CompetingInterests}

The authors declare no competing interests.

\ifproduction

\renewcommand\thesection{Supplementary Section \arabic{section}}
\renewcommand{\thefigure}{Supplementary Figure \arabic{figure}}
\renewcommand{\thealgorithm}{Supplementary Algorithm \arabic{algorithm}}

% Supplementary Sections (in order of appearance in supplementary)
\setcounter{section}{0}
\phantomsection\refstepcounter{section}\label{sec:appendix:assets-description}
\phantomsection\refstepcounter{section}\label{sec:appendix:enviornments-tasks}
\phantomsection\refstepcounter{section}\label{sec:appendix:heat-transfer-semantics-engine}
\phantomsection\refstepcounter{section}\label{sec:physics-semantics-engine-algorithm}
\phantomsection\refstepcounter{section}\label{sec:appendix:chemistry-experiment-workflows}
\phantomsection\refstepcounter{section}\label{sec:appendix:Deployment-Digital-Twin-Workflows-Real-Setup}

% Supplementary Figures (in order of appearance in supplementary)
\setcounter{figure}{0}
\phantomsection\refstepcounter{figure}\label{fig:heat_transfer}
\phantomsection\refstepcounter{figure}\label{fig:additional_flow_chart}
\phantomsection\refstepcounter{figure}\label{fig:organic_workflow_verification_40_degree}
\phantomsection\refstepcounter{figure}\label{fig:wet_lab_exp}

% Supplementary Algorithm (only one in supplementary)
\setcounter{algorithm}{0}
\phantomsection\refstepcounter{algorithm}\label{alg:semantics-physics}

\else

\clearpage
\twocolumn
\appendix

\setcounter{section}{0}
\renewcommand\thesection{Supplementary Section \arabic{section}}

\titleformat{\section}[block]{\normalfont\large\bfseries}{\thesection}{0.5em}{}
\titleformat{\subsection}{\normalfont\normalsize\bfseries\sffamily}{\thesubsection}{0.5em}{\titlecap}
\titleformat{\subsubsection}{\normalfont\normalsize\bfseries\sffamily}{\thesubsubsection}{0.5em}{\titlecap}

\setcounter{figure}{0} \captionsetup[figure]{name={Supplementary Figure}}
\renewcommand{\thefigure}{Supplementary Figure \arabic{figure}}
\renewcommand{\thealgorithm}{Supplementary Algorithm \arabic{algorithm}}

\section{Assets Description}
\label{sec:appendix:assets-description}

\subsection{Materials}
\ourmodel supports solids and fluids using position-based-dynamics (PBD) particle simulation.

\subsection{Glassware}
\ourmodel has different glassware available and readily-used in our environments. These include beakers, graduated cylinders, reagent bottles, funnels, test tubes and vials.

\subsubsection{Laboratory Tools}

\textit{Digital Pipette}: This can be integrated with a robot arm to facilitate the processing of pipetting fluids through its digital interface.

\textit{Spatulas}:  Human scientists weigh samples (materials) using a spatula and balance.
The spatula used in our environment has been demonstrated to autonomously weigh different solid materials in our previous works~\cite{Jiang2023SolidDispensing}.

\subsection{Laboratory Instruments}
\textit{Hot plate, stirrer, and balance}: A key instrument is a hot plate, balance, and stirrer, where for this setup we used the IKA RCT Digital hot plate and stirrer, which has been used in our previous works for solubility screening and crystallisation~\cite{fakhruldeen2022archemist, yoshikawa2023large}.

\textit{Syringe pump}: Syringe pumps are commonly used for precise and reliable liquid handling. In our work, we have used the Tecan Cavro pumps.

\subsection{Automation Platforms}
\textit{Opentrons}: OT-2 is an open-source bench-top liquid handler.
\subsubsection{Robots}
\textit{KUKA KMR iiwa}: is a mobile manipulator with a 7 DoF arm that has been used in different material discovery experiments including the \update{first}{} mobile robotic chemist for photocatalysis~\cite{Burger2020} and the multi-robot system for solid-state chemistry~\cite{Lunt2024}. 

\textit{Franka Emika Panda}: The Franka Emika Panda arm robot, equipped with either a Franka Emika Panda Hand or a Robotiq 2F-85 gripper was used as the main manipulator in the environment since it has already been demonstrated as a robotic chemist across different workflows\cite{darvish2024organa, Pizzuto2022SOLIS}. 
To facilitate grasping objects from the side in tabletop scenarios and enhance constrained motion planning in some of the environments, we positioned the Robotiq 2F-85 end-effector parallel to the ground on the robot’s last link.
This configuration was attained either through a fixed linkage or by incorporating a Dynamixel XM540-W150 servo motor as an additional degree of freedom.

\section{Chemistry Lab Simulated Environments}
\label{sec:appendix:enviornments-tasks}

\update{}{The videos of all the environment are provided on the project website at:\\ \url{https://accelerationconsortium.github.io/Matterix/}.}

\textbf{Beaker Placement in Fumehood with Franka Emika Panda 8 DoF robot arm and Robotiq gripper}: The Franka Emika Panda 8 DoF robot arm, equipped with a Robotiq gripper, is precisely manoeuvring a beaker within the confines of a fumehood. The robot is executing a controlled placement of the beaker, demonstrating its ability to handle laboratory equipment in a potentially hazardous environment.
\update{YouTube link: \url{https://youtu.be/tRQyJFPwgyo} . An example reaching task using the cuRobo motion planner with collision avoidance is demonstrated here:  \url{https://youtu.be/EWxdYJpaqvQ}}{An example reaching task using the cuRobo motion planner with collision avoidance is demonstrated as well.}

\textbf{Beaker Placement in OT2 with Franka Emika Panda 8 DoF robot arm and Robotiq gripper}: The Franka Emika Panda robot with its Robotiq gripper is executing a precise placement of a beaker within the OT2 liquid handling platform's workspace. 
\update{YouTube link: \url{https://youtu.be/V1CARgFN2wQ}}{}

\textbf{{Funnel Pick and Place with Franka Emika Panda 8 DoF robot arm and Robotiq gripper}}: The Franka Emika Panda with its Robotiq gripper, is performing a pick-and-place operation with a funnel. This involves grasping the funnel, transporting it, and accurately positioning it in a new location.

\textbf{{KUKA KMR IIWA Pick and Place Beaker}}: The mobile KUKA KMR IIWA robot is autonomously transporting a beaker between different locations in a representative lab.
\update{YouTube link: \url{https://youtu.be/NHiyla50Who}}{}

\textbf{{Liquid Dispensing Station}}: The robot arm (Franka Emika Panda) sets up the OT-2 workstation by placing the dispensing rack, tips rack, and reservoir well in the tray. Following this, the OT-2 head connects the pipettes and begins the fluid dispensing process.
\update{YouTube link: \url{https://youtu.be/DWISnyqtzUk}}{}

\textbf{Franka Emika Panda (8DoF with Robotiq gripper) Pick and Place Beaker on Scale}: The robot arm is accurately placing a beaker onto the scale, for weighing its contents.

\textbf{Franka Emika Panda (8DoF with Robotiq gripper) Pour Fluid}: The robot is precisely pouring a liquid from one container to another.

\textbf{Franka Emika Panda Pick and Place Vial Rack}: The robot arm is transporting the racks of vials between different locations.

\textbf{Franka Emika Panda (8DoF with Robotiq gripper)/Franka Emika Panda Vial Insertion}: The robot arm is picking up a vial and inserting it in the rack.
\update{YouTube link: \url{https://youtu.be/Y5Sf8-DDKDQ}}{}

\textbf{Franka Emika Panda Wellplate Insertion in OT2}: The robot arm is picking up the wellplate and inserting it in the rack locations on the OT2.

\update{To run the \ourmodel environments listed above, the following command should be used.}{}

\update{Where:}{}

\update{Detailed instructions for running these environments are provided in the README file of the repository. We encourage the reader to review this documentation.
}{Detailed instructions and tutorials for running \ourmodel are available on the project website and repository. We encourage readers to review them.}
\update{The software package and associated assets are available for review in the provided online folder: {\small \url{https://drive.google.com/drive/folders/}} This material is confidential and intended solely for editors and reviewers. Please do not distribute, as we are preparing an official release.}{}

\update{A video tutorial demonstrating \ourmodel and how to run/visualize its environment is available at: \url{https://youtu.be/GcKD29v4hqU}.}{}

\section{Heat Transfer Semantics Engine\update{}{Example}}
\label{sec:appendix:heat-transfer-semantics-engine}

Thermal analysis is crucial in laboratory systems to ensure experimental accuracy and optimize energy efficiency. Here, we develop a thermal model for a solution-beaker-heater system as an example, applying an energy balance approach to predict temperature variations in different components over time. This approach is generic and could be applied to different laboratory components.

\begin{figure}[!h]
\includegraphics[width=0.49\textwidth, trim=0mm 20mm 20mm 20mm, clip]{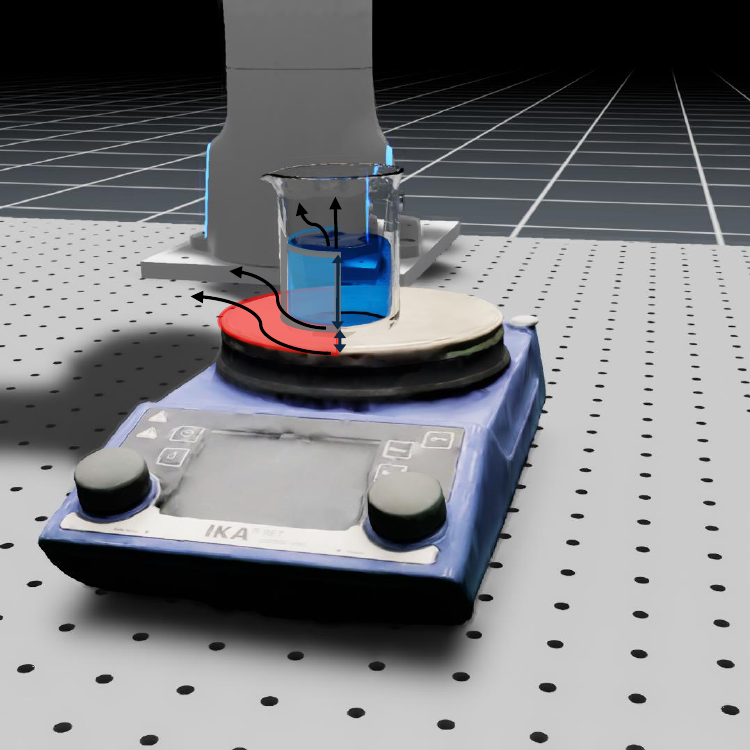}
    \caption{: \textbf{Heat transfer between the heater, beaker, and solution.} The IKA RET control device heats the beaker through conduction, which then transfers heat to the solution. Heat exchange is bidirectional, flowing from hotter to cooler objects until equilibrium is reached. Curved arrows indicate heat dissipation to the ambient air.}
    \label{fig:heat_transfer}
\end{figure}

\paragraph{System Overview}
The system consists of three interconnected thermal components:
\begin{itemize}
    \item \textbf{IKA Heater}: A solid heating element providing heat input.
    \item \textbf{Beaker}: A glass container facilitating heat transfer between the heater and the liquid.
    \item \textbf{Solution}: A liquid medium subjected to thermal interactions with the beaker and ambient air.
\end{itemize}
Heat transfer in the system occurs primarily through conduction and convection.

\paragraph{Heat Transfer Mechanisms}
A uniform temperature distribution within each component is assumed in \ourmodel. Each component exchanges heat through different pathways, as shown in \supfigref{fig:heat_transfer}. Different pathways are described below:
\begin{itemize}
    \item IKA Heater
    \begin{itemize}
        \item Conductive heat transfer to the Beaker: \( \dot{q}_{hb} \)
        \item Convective heat loss to the surrounding Air: \( \dot{q}_{ha} \)
        \item Internal heat generation: \( \dot{q}_{\text{gen}} \)
    \end{itemize}
    \item Beaker
    \begin{itemize}
        \item Conductive heat transfer from the IKA Heater: \( \dot{q}_{hb} \)
        \item Conductive heat transfer to the Solution: \( \dot{q}_{bs} \)
        \item Convective heat loss to the Air: \( \dot{q}_{ba} \)
    \end{itemize}
    \item Solution
    \begin{itemize}
        \item Conductive heat transfer from the Beaker: \( \dot{q}_{bs} \)
        \item Convective heat loss to the Air: \( \dot{q}_{sa} \)
    \end{itemize} 
\end{itemize}

\paragraph{Energy Balance Equations}
Applying the first law of thermodynamics, we obtain the governing equations:

\textbf{IKA Heater}

\begin{equation}
    m_h C_h \frac{dT_h}{dt} = \dot{q}_{\text{gen}} - \dot{q}_{hb} - \dot{q}_{ha}
    \label{eq:energy_balance_heater}
\end{equation}
where:
\begin{equation}
    \dot{q}_{hb} = \frac{k_b A_b}{d_b} (T_h - T_b) \quad
\end{equation}
\begin{equation}
    \dot{q}_{ha} = h_h A_h (T_h - T_a) \quad
\end{equation}
The term \(\dot{q}_{\text{gen}}\) represents the heat generated internally within the IKA heater due to electrical power input. Generally, the heater can be assumed to behave as a purely resistive element, the heat generation follows Joule’s Law (${V^2}/{R} = RI^2$, where $V$ is voltage, $I$ is current, and $R$ is the resistance). However, in our case, we are mainly facing a temperature-controlled heating, where the heater adjusts its heat generation to maintain a target temperature. 
Therefore, instead assuming a constant electrical power supply, \(\dot{q}_{\text{gen}}\) depend on the heater temperature using a proportional control system, where the heater generates heat depending on the difference between its current temperature, ($T_h$), and the target temperature, ($T_{target}$):
\begin{equation}
    \dot{q}_{\text{gen}} = K_{gen}(T_{target}-T_h)
\end{equation}

\autoref{eq:energy_balance_heater} describes the energy balance for the heater, where the generated heat (\(\dot{q}_{\text{gen}}\)) is partially transferred to the beaker via conduction (\(\dot{q}_{hb}\)) and lost to the ambient through convection (\(\dot{q}_{ha}\)).

\textbf{Beaker}

\begin{equation}
    m_b C_b \frac{dT_b}{dt} = \dot{q}_{hb} - \dot{q}_{bs} - \dot{q}_{ba}
\end{equation}
where:
\begin{equation}
    \dot{q}_{bs} = \frac{k_b A_b}{d_b} (T_b - T_s) \quad
\end{equation}
\begin{equation}
    \dot{q}_{ba} = h_b A_b (T_b - T_a) \quad
\end{equation}

\textbf{Solution}
\begin{equation}
    m_s C_s \frac{dT_s}{dt} = \dot{q}_{bs} - \dot{q}_{sa}
\end{equation}
where:
\begin{equation}
    \dot{q}_{sa} = h_s A_s (T_s - T_a) \quad
\end{equation}

\textbf{Parameter Definitions}
\begin{itemize}
    \item \( m \) ($kg$): Mass of the component
    \item \( C \) ($J/kg \cdot K$): Specific heat capacity
    \item \( T \) ($K$): Temperature
    \item \( T_h \) ($K$): Heater temperature
    \item \( T_b \) ($K$): Beaker temperature
    \item \( T_a \) ($K$): Ambient temperature
    \item \( k \) ($W/m \cdot K$): Thermal conductivity
    \item \( A \) ($m^2$): Heat transfer area
    \item \( d \) ($m$): Thickness of conductive medium
    \item \( h \) ($W/m^2 \cdot K$): Convective heat transfer coefficient
    \item \( \dot{q} \) ($W$): Heat transfer rate
    \item \( V \) (V): Voltage applied across the heater
    \item \( R \) (\(\Omega\)): Electrical resistance of the heater
    \item \( I \) (A): Current flowing through the heater
    \item \( K_{gen} \) ($W/K$): Proportional control gain, which determines how aggressively the heater responds to temperature differences
    \item \( T_{target} \) ($K$): Target temperature of the IKA heater
\end{itemize}

\paragraph{Uniform Temperature Distribution Assumption}

The model employed in this work assumes a uniform temperature distribution within each component, meaning that no internal temperature gradients exist. 
This assumption is valid when the Biot number ($Bi$) is sufficiently small, indicating that internal conductive resistance is much lower than surface convective resistance~\cite{lienhard2008jh}. 
The Biot number is defined as: 

\begin{equation}
    \label{eq:biot}
    \mathrm{Bi}=\frac{h L_c}{k}
\end{equation}
where, $h$ is the convective heat transfer coefficient of the surrounding fluid ($W/m^2 \cdot K$) , \( k \) is the thermal conductivity of the material ($W/m\cdot K$), and \( L_c \) is the characteristic length ($m$), given by: 
\begin{equation}
    \label{eq:lc}
    L_c = \frac{\mathcal{V}}{A}
\end{equation}
where, \( \mathcal{V} \) and \( A \) represent the volume and  surface area of the object, respectively.  

Although the Biot number in this study may exceed the typical threshold ($0.1$) depending on the object carried by the robotic system, assuming a uniform temperature distribution remains a practical simplification. 
This approach allows for an efficient thermal analysis that can be integrated into the \ourmodel to predict the temperature variations effectively.

In addition, it is worth mentioning that to determine whether heat transfer inside the solution is dominated by convection or conduction, we can use the Rayleigh number ($Ra$)~\cite{lienhard2008jh, kerr1996rayleigh}. The Rayleigh number helps predict whether buoyancy-driven fluid motion (natural convection) occurs. It is defined as follow: 
\begin{equation}
    \label{eq:ra}
    Ra = Gr \cdot Pr
\end{equation}
where:
\begin{itemize}
    \item \( Gr \) is the Grashof number, representing the ratio of buoyancy to viscous forces.
    \item \( Pr \) is the Prandtl number, representing the ratio of momentum diffusivity to thermal diffusivity.
\end{itemize}
The Grashof number is given by:
\[
Gr = \frac{g \beta (T_{\text{hot}} - T_{\text{cold}}) L^3}{\nu^2}
\]
where:
\begin{itemize}
    \item \( g \) = gravitational acceleration (\(9.81\) m/s²),
    \item \( \beta \) = thermal expansion coefficient of the liquid (\(1/K\)),
    \item \( T_{\text{hot}} - T_{\text{cold}} \) = temperature difference in the liquid,
    \item \( L \) = characteristic length (\update{e.g.}{for example}, height of the liquid column),
    \item \( \nu \) = kinematic viscosity of the liquid (\(m^2/s\)).
\end{itemize}
The Prandtl number is given by:
\[
Pr = \frac{\nu}{\alpha}
\]
where:
\begin{itemize}
    \item \( \alpha \) = thermal diffusivity (\(m^2/s\)),
    \item \( \nu \) = kinematic viscosity (\(m^2/s\)).
\end{itemize}

The interpretation of the Rayleigh number is as follow:
\begin{itemize}
    \item If \( Ra \gg 10^4 \) (high \( Ra \)) \( \Rightarrow \) Convection dominates (buoyancy-driven fluid motion).
    \item If \( Ra \ll 10^4 \) (low \( Ra \)) \( \Rightarrow \) Conduction dominates (the liquid remains mostly still).
    \item If \( Ra \approx 10^4 \) \( \Rightarrow \) Transition region where both conduction and convection occur.
\end{itemize}

This work establishes an energy balance framework for multi-object systems, providing heat transfer equations for analysis and simulation integrated with chemistry and object manipulation. Future work may involve online identification of heat transfer constants using IoT sensors to align simulations with real-world results.

\section{Physics and Semantics Engine Algorithm}
\label{sec:physics-semantics-engine-algorithm}

\ref{alg:semantics-physics} describes the integration of the Physics and Semantics engines. 
At each simulation step, we first pass the current physical states $\bm{x}_{t}$ and actions $\bm{a}_{t}$ to the physics engine to compute the next physics state. Next, we iterate over all the semantics processes and events to estimate the next semantic states $\bm{s}_{t+1}$ and $\bm{l}_{t+1}$. Additionally, events can influence the behaviours of both the physics and semantics.

\begin{algorithm}[H]
\scriptsize
\begin{algorithmic}[1]
\captionsetup{justification=raggedright, singlelinecheck=false}
\caption{\textsc{Physics\&SemanticsEngine()}}
\label{alg:semantics-physics}
\Require {$\bm{x}_{0},\bm{s}_{0},\bm{l}_{0}, \bm{a}_{t} $ }
\Ensure {$\bm{x}_{t+1},\bm{s}_{t+1},\bm{l}_{t+1}$}
\ForAll{$t \in \mathcal{T}$} 
\State $\bm{x}_{t+1} \leftarrow$  $\bm{F}_{t}$($\bm{x}_{t}, \bm{a}_{t}$) \Comment{physics engine}
\State $\bm{s}_{t+1} \leftarrow$  $\bm{P}_{t}$($\bm{x}_{t+1}, \bm{s}_{t}, \bm{l}_{t}, \bm{a}_{t}$) \Comment{semantic processes}
\State $\bm{l}_{t+1}, \bm{F}_{t+1}, \bm{P}_{t+1} \leftarrow$  $\bm{E}$($\bm{x}_{t+1}, \bm{s}_{t+1}, \bm{l}_{t}, \bm{a}_{t}$) \Comment{semantic events}
\EndFor
\end{algorithmic}
\end{algorithm}

\section{Chemistry Experiment Workflows}
\label{sec:appendix:chemistry-experiment-workflows}

\revise{}{\supfigref{fig:additional_flow_chart} illustrates the hierarchical state machine used to run multi-scale chemistry simulations. \supfigref{fig:additional_flow_chart}.A shows the state machine that executes the organic chemistry experiment presented in \autoref{fig:workflow_verification}.A, controlling both the heater and the robot.}
\revise{}{\supfigref{fig:additional_flow_chart}.B shows the state machine for the two-step reduction-oxidation chemistry experiment presented in \autoref{fig:workflow_verification}.B.
The robot begins by pouring the solution from the first beaker into the third, followed by the solution from the second beaker, mixing all three solutions in the third beaker. In both workflows, the robot’s motion is calculated using inverse kinematics.}

\revise{}{
\supfigref{fig:organic_workflow_verification_40_degree} presents the results of the organic chemistry experiment conducted with the IKA plate heater set to a target temperature of 40$^\circ\text{C}$. As shown in the figure, the solution temperature remains close to 35$^\circ\text{C}$, resulting in a relatively slow rate for the organic reaction.
}

\begin{figure*}[!th]
    \centering 
    \includegraphics[width=0.95\textwidth]{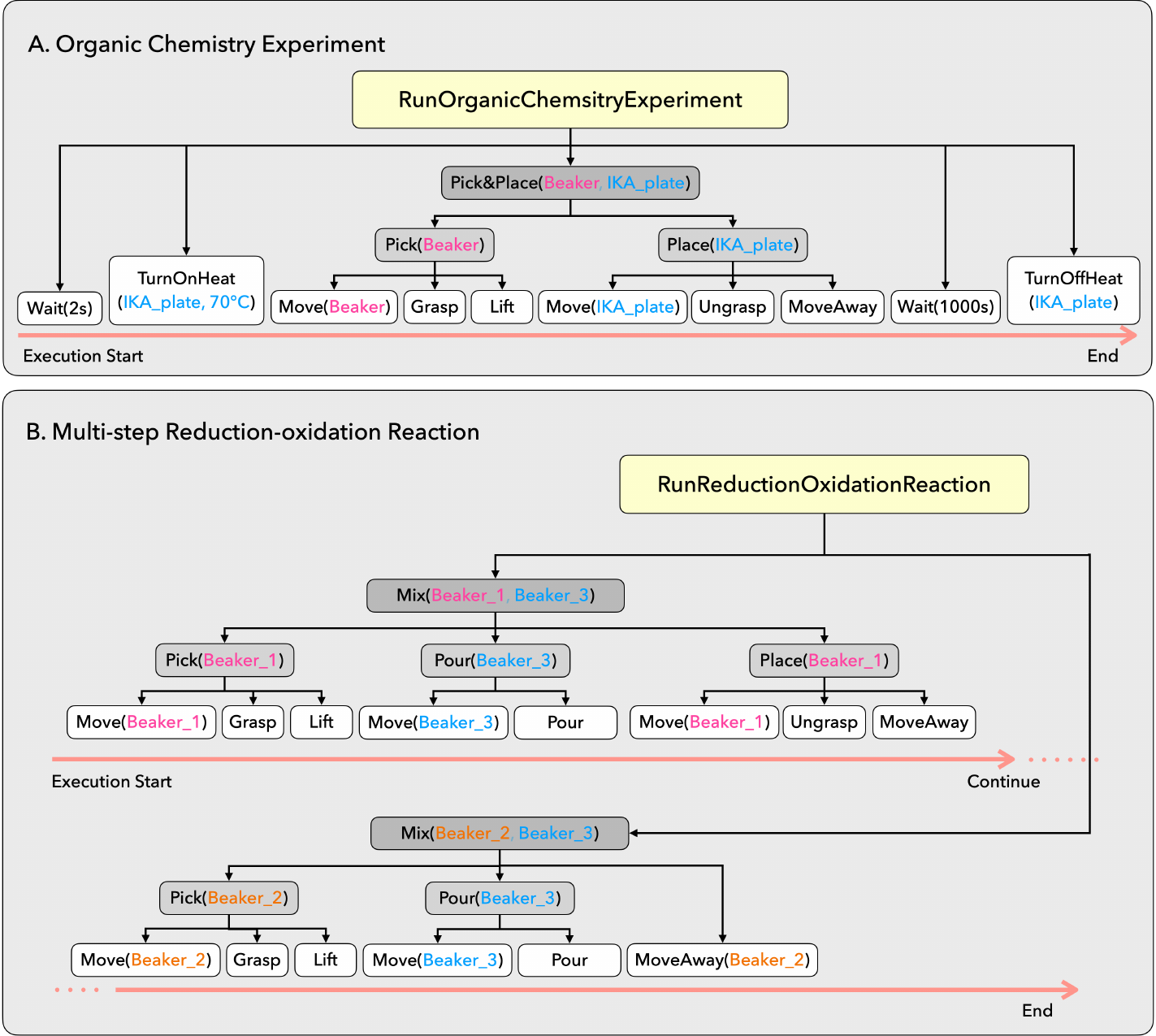}
    \caption{: \revise{}{\textbf{Visualization of hierarchical state machine for different chemistry experiments.} (A) demonstrates the workflow of the organic chemistry experiment as shown in \autoref{fig:workflow_verification}.A. (B) demonstrates the workflow of the multi-step reduction-oxidation reaction as shown in \autoref{fig:workflow_verification}.B.  }}
    \label{fig:additional_flow_chart}
\end{figure*}

\begin{figure*}[!th]
    \centering 
    \includegraphics[width=0.95\textwidth]{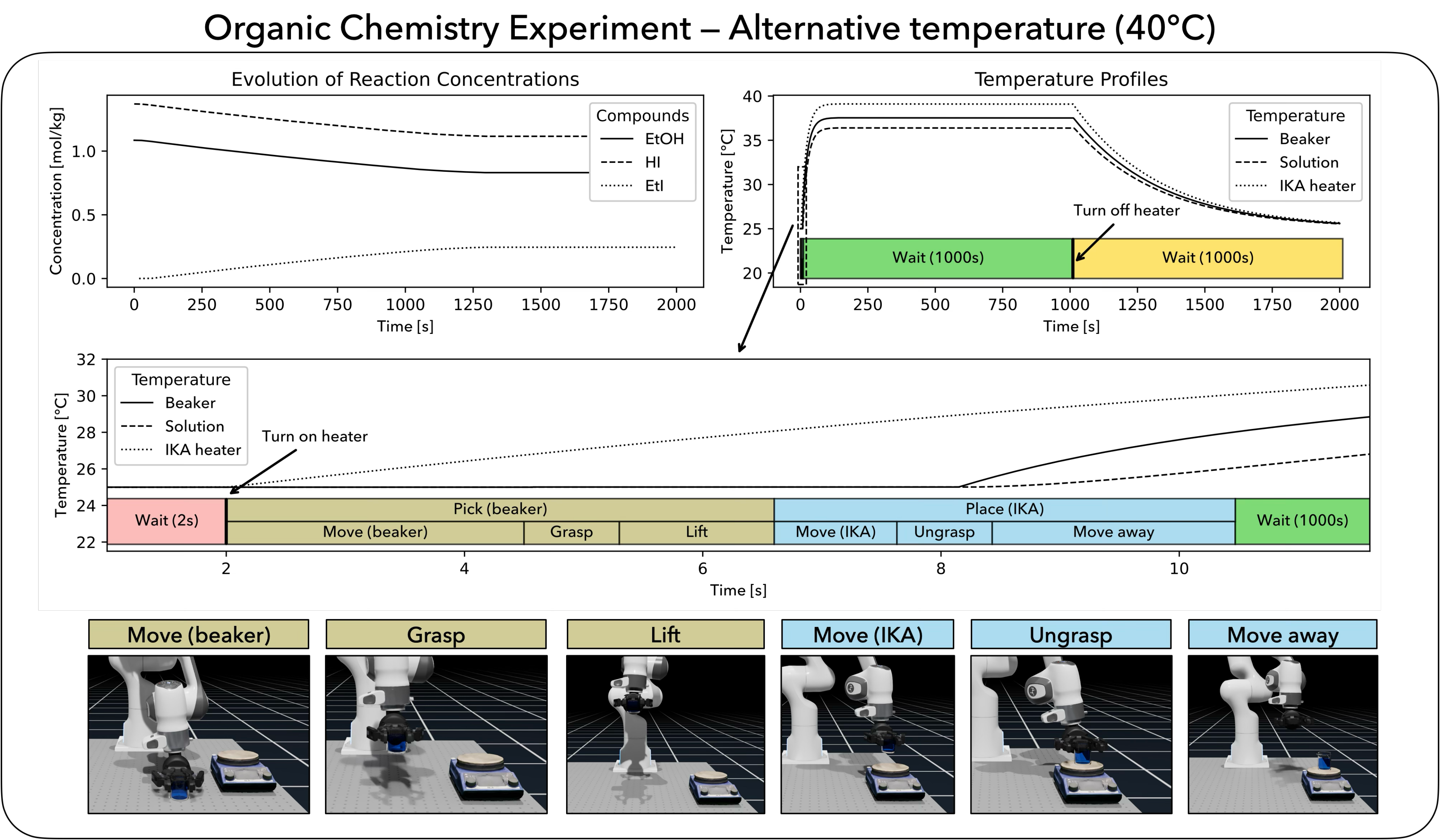}
    \caption{: \revise{}{\textbf{Multi-scale simulation of an organic chemistry experiment.} A single-step organic chemistry experiment demonstrating the interaction of physical manipulation, heat transfer, and chemical kinetics simulation. The target temperature for the IKA plate heater is set to 40$^\circ\text{C}$. Compared to~\autoref{fig:workflow_verification}.A, the reaction proceeds at a slower rate at 40$^\circ\text{C}$.
    }}
    \label{fig:organic_workflow_verification_40_degree}
\end{figure*}

\paragraph{\update{Reduction-Oxidation Execution Protocol}{}}
\update{The following stock solutions were prepared manually using an analytical balance, all chemical were purchased from Sigma Aldrich.\newline}{}
\update{\textbf{Solution 0:} $\mathrm{KI}$ (0.66 g, 4 mmol) and saturated starch in 100 g $\mathrm{H_2O}$, total concentration of $\mathrm{KI}$ = 0.04 mol/kg\newline}{}
\update{\textbf{Solution 1:} $\mathrm{H_2O_2}$ (3 g of 5\% w/w, 4.4 mmol) in 100 g $\mathrm{H_2O}$, total concentration of $\mathrm{H_2O_2}$ = 0.044 mol/kg\newline}{}
\update{\textbf{Solution 2:} ascorbic acid (0.88 g, 5 mmol) in 100 g $\mathrm{H_2O}$, total concentration of ascorbic acid = 0.05 mmol/kg\newline}{}
\update{\textbf{Step 1:}}{}
\update{To a large reaction beaker (\textbf{Beaker 0}) containing 100 g of \textbf{Solution 0} (4 mmol $\mathrm{I^-}$, saturated starch) over an analytical balance, the robotic arm picked up \textbf{Beaker 1} containing \textbf{Solution 1} and poured around 50 g \textbf{Solution 1} into the reaction beaker. An increase in total mass was measured to be 48 g, total amount of $\mathrm{H_2O_2}$ added was calculated to be 2.1 mmol.\newline}{}
\update{The mixture was left for reaction for 50 s, and the color gradually turned dark blue due to the oxidation of $\mathrm{I^-}$ into $\mathrm{I_2}$ which immediately formed a dark blue-brown iodine-starch complex.\newline}{}
\update{\textbf{Step 2:}}{}
\update{To the reaction beaker containing product of \textbf{Step 1} , the robotic arm picked up \textbf{Beaker 1} containing \textbf{Solution 2} and poured around 50 g \textbf{Solution 2} into \textbf{Beaker 0}. An increase in total mass was measured to be 48 g, total amount of ascorbic acid added was calculated to be 2.4 mmol.\newline}{}
\update{The mixture was left for reaction for 50 s, and the dark blue-brown color gradually faded due to the reduction of $\mathrm{I_2}$ into $\mathrm{I^-}$, which lead to the subsequent consumption of the iodine-starch complex.\newline}{}

\begin{figure*}[!th]
    \centering 
    \includegraphics[width=0.95\textwidth]{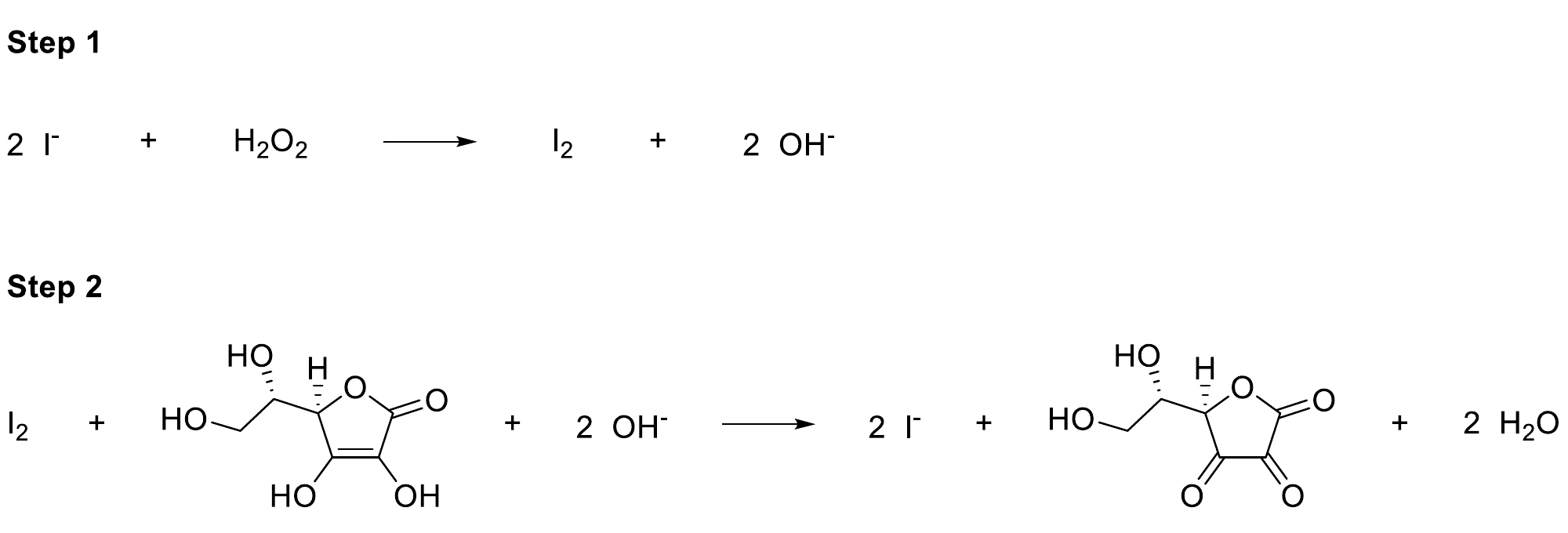}
    \caption{: \textbf{Reaction equations of two step organic-inorganic redox-oxidation experiment.} \update{}{(Step 1) Oxidation of iodide to elementary iodine with hydrogen peroxide, reflected by the dark color of the iodine-starch complex. (Step 2) Reduction of iodine with ascorbic acid (Vitamin C), reflected by the diminishing of the iodine-starch dark color.}}
    \label{fig:wet_lab_exp}
\end{figure*}

\section{Deployment of Digital Twin Workflows to Real Setup }
\label{sec:appendix:Deployment-Digital-Twin-Workflows-Real-Setup}
\update{}{The videos of the deployment of digital twin workflows to the real setup are provided on the project website at: \\
\url{https://accelerationconsortium.github.io/Matterix/}.}

\textbf{Robot Pouring Liquid from a beaker to another}: 
The Franka Emika Panda 7 DoF robot arm, equipped with a Robotiq gripper, is transfers liquids during a pouring task (related to \autoref{fig:task_deployment}.B).
\update{YouTube link: \url{https://youtu.be/vb9vNfW8HJ4}}{}

\textbf{Liquid Hnadling Station with Opentrons and Franka Arm}: 
The Franka Research 3 (FR3) robot arm setups the Opentrons liquid hanlder in a real-world chemistry lab (related to \autoref{fig:task_deployment}.C).
\update{YouTube link: \url{https://youtu.be/DtlD9_smsqk}}{}

\textbf{Multi-step Reduction-Oxidation Chemistry Experiment}: 
The Franka Panda arm (FR3) robot arm  equipped with a Robotiq gripper runs a two-step reduction-oxidation chemistry experiment (related to \autoref{fig:workflow_verification}.B).
\update{YouTube link: \url{https://youtu.be/11p_eiRAjns}}{}

\fi

\end{document}